\documentclass[11pt]{article}

\usepackage[final]{acl}
\usepackage{enumitem}
\usepackage{times}
\usepackage{latexsym}
\usepackage{hyperref}
\usepackage[T1]{fontenc}

\usepackage[utf8]{inputenc}
\usepackage{fvextra}

\usepackage{microtype}

\usepackage{inconsolata}

\usepackage{booktabs}
\usepackage{amsmath}
\usepackage{graphicx}
\usepackage[normalem]{ulem}
\usepackage{amssymb}
\usepackage{mdframed}
\definecolor{promptbg}{RGB}{245,245,245}
\definecolor{promptpanelbg}{RGB}{248,250,252}
\definecolor{promptpanelhead}{RGB}{226,232,240}
\definecolor{promptpanelborder}{RGB}{148,163,184}
\definecolor{promptpaneltext}{RGB}{30,41,59}

\newenvironment{promptpanel}[2]{%
  \begin{minipage}{0.96\textwidth}
  \small
  \hrule height \heavyrulewidth
  \vspace{0.55ex}
  \noindent\textbf{#1}\hfill\textit{\footnotesize #2}
  \par\vspace{0.45ex}
  \hrule height \lightrulewidth
  \vspace{0.55ex}
}{%
  \vspace{0.45ex}
  \hrule height \heavyrulewidth
  \end{minipage}
}
\newcommand{\promptcaptionsetup}{}

\usepackage{colortbl}
\usepackage{xcolor}
\usepackage[normalem]{ulem}
\usepackage{multirow}
\definecolor{perpcolor}{RGB}{227,242,253}    
\definecolor{clscolor}{RGB}{232,245,233}     
\definecolor{syncolor}{RGB}{255,243,224}      
\definecolor{ourscolor}{RGB}{243,229,245}     

\title{Every Time I Hire a Linguist, Inference Costs Go Down:\\ On Linguistic Rules as Effective Prompt Compressors}

\author{
  \textbf{Jianfei Ma} \quad
  \textbf{Zhaoxin Feng} \quad
  $\textbf{Emmanuele Chersoni}^{\spadesuit}$ \quad
  $\textbf{Si Chen}^{\spadesuit}$ \\
  The Hong Kong Polytechnic University \\
  \texttt{\{jianfei-mark.ma,zhaoxinbetty.feng\}@connect.polyu.hk} \\
  \texttt{\{emmanuele.chersoni,sarah.chen\}@polyu.edu.hk}
}

\begin{document}
\maketitle

\begin{abstract}
Prompt compression shortens LLM input to reduce inference cost, yet existing methods score token importance through LM forward passes. It remains questionable whether such nuanced, costly token selection is necessary. Compression requires identifying informative content, a problem that linguistic research has long addressed through cues that can be operationalized as deterministic rules. We therefore ask: can \textbf{linguistic rules alone} serve as effective prompt compressors, without LM-based scoring at compression time?


To address this, we conduct offline evolutionary search over lexical, syntactic, semantic, and discourse seeds to find competitive rule combinations. The resulting linguistic compressor requires no LM forward pass at deployment and uses only CPU-side processing for compression. We evaluate it with a dual-path protocol to balance compression quality and reconstruction fidelity.

Across short passages, multi-document reasoning, and dialogue-memory QA datasets, evolved compressors achieve performance similar to that of recent advanced prompt-compression strategies. Performance is strongest under light-to-moderate compression and degrades as compression becomes more aggressive, while the Direct and Reconstruction paths exhibit distinct patterns. Evolutionary analysis reveals that effective compression fuses signals across linguistic levels and, as the compression ratio increases, rules shift from token pruning to sentence extraction.\footnotetext{$\spadesuit$ represents corresponding authors.} \footnote{Code and data are available at \url{https://github.com/sdmjf/LingComp}.}
\end{abstract}

\section{Introduction}

Large language models (LLMs) are increasingly deployed in long-context applications like retrieval-augmented generation (RAG)~\cite{lewis2020retrieval}, multi-agent communication~\cite{li2023camel}, and long-term dialogue~\cite{wu2025longmemeval}, where massive contextual data often overshadows the core task instruction. This drives up inference costs, increases latency, and degrades model comprehension by diluting key information. Hard prompt compression~\cite{li-etal-2025-prompt}, which shortens the input by pruning tokens while preserving natural language readability, has emerged as a standard remedy. Yet the dominant paradigm relies on language model (LM) scoring, whether via perplexity or trained classifiers, to assess token importance~\cite{jiang-etal-2023-llmlingua,li-etal-2023-compressing,pan-etal-2024-llmlingua,raiyan2025frugalpromptreducingcontextualoverhead}. This raises two critical concerns: (1) such scoring is computationally expensive, forward pass-dependent, and inherently opaque; (2) token-level granularity risks severing syntactic dependencies and semantic coherence. Given such drawbacks, a natural question arises: is fine-grained, per-token model scoring the only path to effective compression?

\begin{figure}[t]
  \includegraphics[width=\columnwidth]{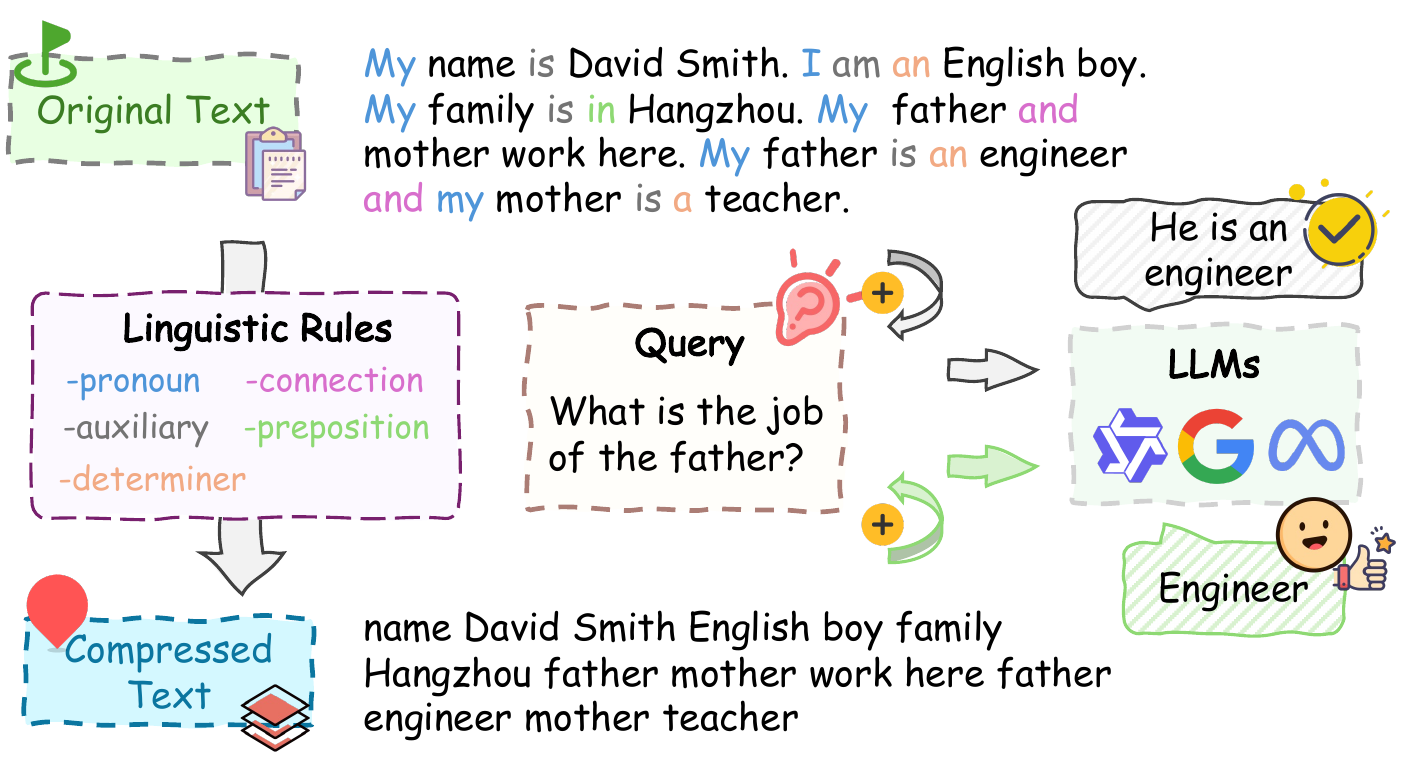}
  \caption{Illustration of linguistic rule-based prompt compression. By following specific linguistic rules, the original text is transformed into a compressed form. LLMs can still correctly answer the same query from both the original and compressed inputs.}
  \label{fig:experiments}
\end{figure}

Linguistics has long studied these signals: content-versus-function word distinctions at the lexical level~\cite{shannon1948mathematical,zipf2016human}, argument structure at the syntactic level~\cite{tesniere2015elements}, negation and quantifier scope at the semantic level~\cite{horn1989negation,barwise-cooper-1981}, and topic sentence placement and entity coherence at the discourse level~\cite{halliday2014cohesion}. These can be extracted by standard NLP tools at low cost, and each is an explicit, inspectable property rather than a vague score directly generated by models. Furthermore, by operating over structural units such as phrases and clauses, linguistic analysis can better preserve contextual integrity. Yet linguistic rules have never been systematically explored for prompt compression. Even PartPrompt~\cite{Mao2026-px}, which builds compression on syntactic parse trees, operates at a single linguistic level. We therefore pose a fundamental question: \textit{can linguistic rules alone serve as effective prompt compressors?}

However, these rules originate from distinct subfields and research traditions. Neither a unified inventory nor a principled approach to their composition exists, and the cross-level combinatorial space is prohibitively large for manual search.

We address the issue via an offline evolutionary program search. It starts from 42 seeds that encode explicit lexical, syntactic, semantic, and discourse rules, alongside composite pipelines chaining these rules. We conduct the search with OpenEvolve,\footnote{\href{https://github.com/algorithmicsuperintelligence/openevolve}{OpenEvolve} is an open-source implementation of AlphaEvolve~\cite{novikov2025alphaevolvecodingagentscientific}.} using LLMs as mutators to iteratively refine rules and discover novel cross-level combinations. The search converges on an evolved linguistic rule-based compressor, implemented as a Python program. During deployment, it relies exclusively on the standard spaCy pipeline~\cite{honnibal2020spacy} and requires no LM forward pass.


We evaluate across four datasets encompassing short passages, multi-document reasoning, and dialogue memory, via a dual-path protocol testing both directly compressed and reconstruction-based QA. In many settings involving medium-to-long documents, our evolved compressors achieve performance similar to that of recent advanced prompt-compression strategies, without an LM forward pass at deployment and with substantially lower computational costs. Across three receivers, they achieve the lowest variability on RACE, remain among the most stable methods on Multi-doc QA, and are competitive on Qasper and LongMemEval. Additionally, our evolutionary analysis shows that the best compressors found by our search consistently fuse signals across multiple linguistic levels, while the dominant compression unit shifts from token-level pruning to sentence-level extraction as budgets tighten. In summary, our work makes three contributions:
\begin{itemize}[itemsep=0pt, topsep=1pt, parsep=1pt]
    \item \textbf{Framework.} To our knowledge, we introduce one of the first prompt compression frameworks centered on explicit linguistic rules and evolved via LLM-guided search.
    \item \textbf{Empirical findings.} We show that rule-based compressors can achieve performance within a range similar to that of recent advanced prompt-compression strategies in many longer-document settings, with low cross-model variability on two datasets and computationally efficient deployment.
    \item \textbf{Analytical insights.} Effective compression fuses multiple linguistic levels, and the optimal granularity shifts from tokens to sentences as compression deepens.
\end{itemize}

\section{Related Work}
\subsection{Prompt compression}

Prompt compression reduces input length to lower LLM inference cost while preserving task-relevant
information~\cite{nagle2024fundamental,zheng2025an,li-etal-2025-prompt,10.1007/978-3-032-21289-4_17}. \textit{Soft prompt methods}~\cite{mu2023learning,chevalier-etal-2023-adapting,ge2024incontext,li-etal-2025-500xcompressor,cheng2024xrag} can reach much higher ratios through model-specific latent representations, but their outputs are neither readable nor reusable across receiver LLMs. Their ratios and our dual-path protocol are not directly comparable, so our empirical comparison uses hard, readable-text compression.

\begin{figure*}[htbp]
    \centering
  \includegraphics[width=0.95\linewidth]{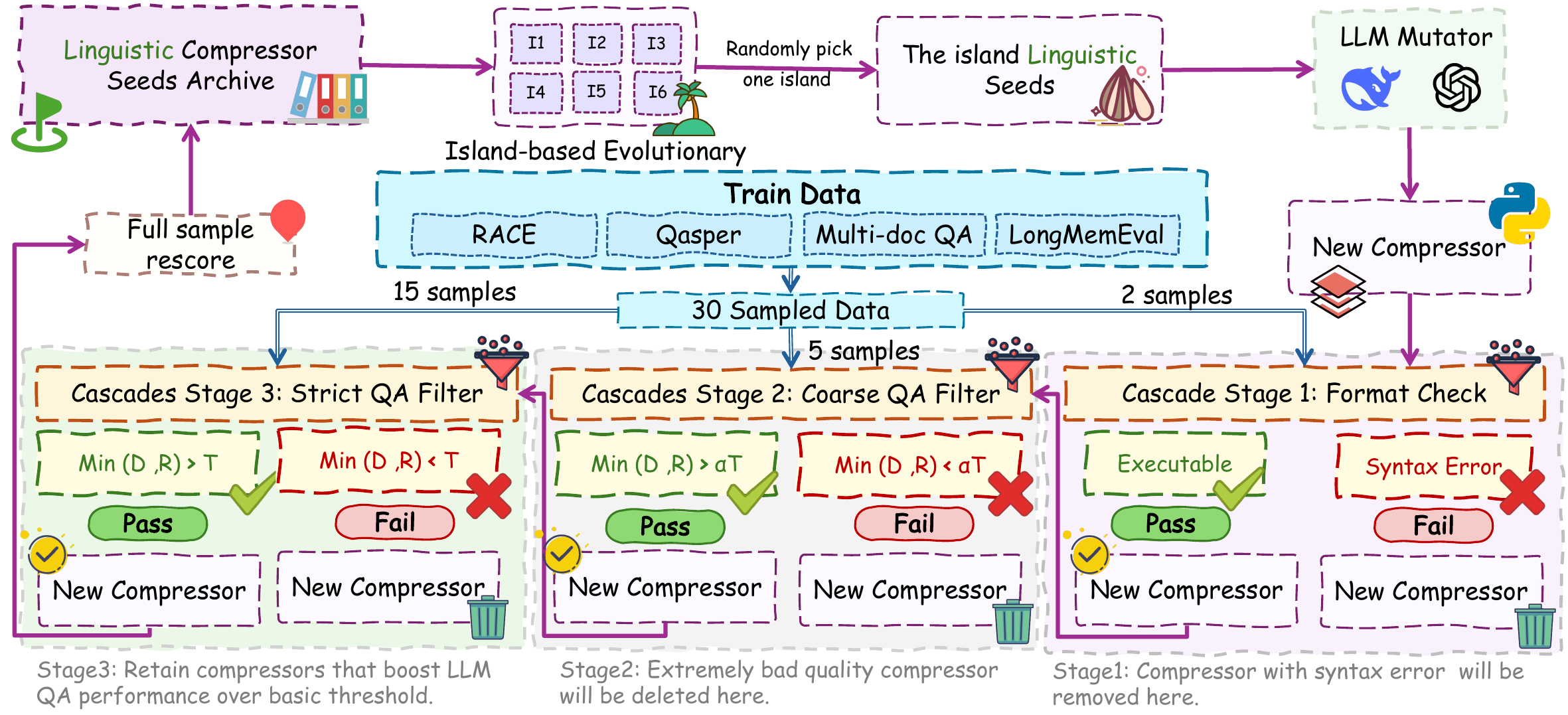}
\caption{Overview of our evolutionary framework for discovering a linguistic prompt compressor. An LLM mutator edits seeds drawn from six
islands, and each new compressor is screened by a three-stage cascade: format check (Stage~1), coarse
QA filter (Stage~2), and strict QA filter
(Stage~3). Survivors are rescored on the full sample and returned to the archive. $T$ is the adaptive acceptance threshold (\ref{sec:Cascaded evaluation}), $\alpha
< 1$ relaxes it at Stage~2, and $\min(D, R)$ is the smaller value of the Direct and Reconstruction QA accuracies (\ref{sec:dual-eval}).}
\label{fig:pipeline}
\end{figure*}

\textit{Hard prompt methods} produce readable compressed text by pruning or rewriting tokens. Selective
Context~\cite{li-etal-2023-compressing} pioneered this direction by estimating token self-information via a causal LM to drop low-information units. Building on this, LLMLingua~\cite{jiang-etal-2023-llmlingua} added coarse-to-fine budget control with iterative token-level pruning, and
LongLLMLingua~\cite{jiang-etal-2024-longllmlingua} extended it to long contexts. LLMLingua-2 \cite{pan-etal-2024-llmlingua} further reframed compression as token classification with a distilled encoder. Selection-p~\cite{chung-etal-2024-selection} replaced distillation with self-supervised continual pre-training for transferability.  KIComp~\cite{LIN2025127738} and FrugalPrompt~\cite{raiyan2025frugalpromptreducingcontextualoverhead} estimate token saliency using attention-based signals, whereas R2C~\cite{choi-etal-2024-reading} ranks chunks and sentences via decoder cross-attention, and ProCut~\cite{xu-etal-2025-procut} estimates attribution over semantic units. A parallel line explores query-conditioned compression, in which retained content adapts to the downstream task~\cite{xu2024recomp,Jung2024-hr,chuang-etal-2024-learning,lajewska-etal-2025-understanding,10.1609/aaai.v39i23.34639}. Despite this diversity, they still need an LM forward pass to judge token importance.

Notable exceptions are PartPrompt~\cite{Mao2026-px} and Telegraph English~\cite{arbuzov2026telegraphenglishsemanticprompt}, which respectively leverage syntactic parse trees and symbolic replacement. However, the former still uses LM-computed scores for node selection, while the latter assumes reliable restoration of manually designed symbols. To avoid both LM scoring and extensive manual design, we use evolutionary search to navigate the multi-level linguistic rule space and construct compressors using only lightweight NLP pipelines.

\subsection{Linguistic Rules for Compact Expression}
\label{sec:related work ling}

Linguistic theories capture language regularities
through compact rules: 
for example, grammar reduces clauses to head-dependent
relations~\cite{tesniere2015elements,de-marneffe-etal-2021-universal}; and rhetorical structure theory abstracts discourse
into a small set of relations~\cite{william1988rhetorical}.
By separating essential structure from 
content, they align naturally with the goals of text compression.



Traditional NLP studies have validated this across distinct levels. At the \textbf{lexical} level, foundational research demonstrates that content words carry propositional meaning while functional elements can be safely discarded for summarization~\cite{shannon1948mathematical,zipf2016human,luhn1958automatic,edmundson1969new}. At the \textbf{syntactic} and \textbf{semantic} levels, compression techniques successfully preserve essential propositions by retaining core argument structures~\cite{KNIGHT200291,filippova-strube-2008-dependency,10.1145/1409360.1409378,yoshikawa-etal-2012-sentence,10.1007/978-3-031-20865-2_20}. Furthermore, \textbf{discourse} connectives and entity chains serve as vital anchors for maintaining coherence across sentences~\cite{prasad-etal-2008-penn,halliday2014cohesion}. These established principles motivate the hypothesis that explicit linguistic rules may support an effective compressor framework.

\subsection{Evolutionary Prompt Optimization}

LLM-guided evolutionary search has proven highly effective at finding optimal solutions in vast discrete search spaces. While current methods like EvoPrompt~\cite{guo2024connecting}, OPRO~\cite{yang2024large}, and PromptBreeder~\cite{10.5555/3692070.3692611} successfully apply this capability to refine natural language instructions, they leave its potential for structural program synthesis unexplored. Constructing a standalone prompt compressor from multidimensional linguistic rules presents an equally massive discrete search challenge. We therefore apply a similar LLM-guided evolutionary paradigm to navigate the linguistic rule space and automatically discover an efficient compression pipeline.

\section{Methodology}


To investigate whether deterministic linguistic
rules can serve as effective prompt compressors, we
design an offline evolutionary search framework
based on AlphaEvolve. The framework has three stages: (1) linguistically motivated seeds form the initial population; (2) offline LLM mutation operators refine compressor code through population-based search guided by compression ratio and downstream QA accuracy; and (3) the top compressors are evaluated on test sets with the dual-path protocol in Figure~\ref{fig:pipeline}.

\subsection{Seed Construction}
\label{sec:seed}

To establish the foundational population for our evolutionary search, we design 42 seeds. Guided by the linguistic literature in \ref{sec:related work ling}, these seeds comprehensively span lexical, syntactic, semantic, and discourse levels, while also including several composite pipelines. Each seed encodes explicit, deterministic rules for identifying informative content. Full specifications are detailed in Appendix~\ref{sec:appendix-seeds}.

\textbf{\textit{10 lexical seeds}} operate at the word level, selecting or removing tokens based on part-of-speech categories, named entity types, and word frequency. Their core strategy is to retain content words and named entities while dropping predictable function words, with additional operations such as short-synonym substitution and symbolic abbreviation.

\textbf{\textit{12 syntactic seeds}} operate within the sentence, exploiting dependency structure to extract core argument triples and prune dispensable subtrees such as relative clauses, coordinated conjuncts, and adjectival or adverbial modifiers.

\textbf{\textit{6 semantic seeds}} preserve meaning-critical markers regardless of compression budget, including negation markers whose removal reverses polarity, modal verbs that encode certainty or obligation, quantifiers that constrain proposition scope, and comparative constructions.

\textbf{\textit{9 discourse seeds}} operate across sentences, using signals such as sentence position, discourse connectives, entity chains, and graph-based sentence ranking to select informative sentences and remove near-duplicate content.

\textbf{\textit{5 composite pipelines}} cascade rules from multiple levels, chaining sentence-level
ranking, dependency-core extraction, SVO triple construction, and symbolic substitution into multi-stage compression sequences.

Regarding implementation, spaCy\footnote{We use \texttt{en\_core\_web\_sm} for linguistic analysis.} is the sole linguistic analyzer used for tokenization, POS tagging, lemmatization, dependency parsing, named entity recognition, and noun chunking.
Some seeds additionally use NLTK~\cite{bird-loper-2004-nltk} with WordNet~\cite{miller-1992-wordnet} for synonym lookup, wordfreq for lexical frequency, and standard Python libraries (\texttt{re}, \texttt{math}, \texttt{collections},
\texttt{numpy}). No LM inference is involved at compression time.

\subsection{Evolutionary Search}
\label{sec:evolve}


\textbf{Framework.} AlphaEvolve~\cite{novikov2025alphaevolvecodingagentscientific} is an evolutionary coding agent that iteratively improves programs via LLM mutation and automated evaluation. Originally built for algorithmic tasks with explicit, stable metrics, we adapt it for prompt compression, as illustrated below.

\textbf{Population and Archive.} Every candidate in our population is a Python script exposing a \texttt{compress(text: str) -> str} function. We divide the population into six independent islands. We distribute the 42 seed compressors evenly across these islands. Each island receives strategies from all four linguistic levels to guarantee initial diversity. In addition, each island manages a local MAP-Elites~\cite{mouret2015illuminatingsearchspacesmapping} archive that categorizes candidates into 48 discrete cells based on code complexity and compression ratio. Every cell keeps only the single highest-scoring compressor. A new candidate can replace the current occupant only if it achieves a higher fitness.

\textbf{Mutation.}
At each iteration, a parent compressor is selected from the archive along with its evaluation feedback, including QA accuracy, compression ratio, and failure examples. An LLM reads the source code and context, then proposes edits to designated mutable regions, which are marked in the source code as evolution blocks. Edits are expressed as search-and-replace pairs, allowing the LLM to adjust scoring weights and filtering conditions, or restructure the compression logic. We use a three-tier ensemble of LLM mutators.



\textbf{Cascaded evaluation.}
\label{sec:Cascaded evaluation}
Full-sample evaluation at every iteration is computationally prohibitive. We therefore design a cascaded pipeline to filter candidates, using increasingly larger sample sizes and stricter thresholds. Only the most promising candidates reach the final full-sample evaluation.
\begin{itemize}[itemsep=1pt, topsep=1pt, parsep=1pt]

\item \textbf{Stage 1.} The compressor runs on two samples with no LLM calls. Candidates are rejected if the output is empty, non-string, or outside the target compression window.

\item \textbf{Stage 2.} Surviving candidates are evaluated on $1/6$ of the samples using
both the Direct and Reconstruction accuracy metrics. A candidate is eliminated if the smaller of its two
path accuracies falls below a threshold, defined as the average no-compression accuracy divided by the target compression ratio and scaled by a constant coefficient smaller than 1.

\item \textbf{Stage 3.} The evaluation sample size is expanded to $1/2$ of the sample pool, and the
screening threshold is tightened to the average no-compression accuracy divided by the target
compression ratio.
\end{itemize}

Candidates passing all three stages are re-evaluated on the full training set to obtain reliable Direct and Reconstruction accuracies, which are used to compute the fitness function and rank compressors within the population.

\textbf{Island Migration.} 
The six islands run their own mutation and selection loops in parallel, exploring different regions of the compressor space, and are connected only through periodic migration under a ring topology~\cite{whitley1999island}. At fixed intervals, the top-fitness fraction of each island is copied to its two adjacent islands, allowing strong candidates to circulate without collapsing all islands into a single population.






\subsection{Dual-Path Evaluation and Fitness}
\label{sec:dual-eval}

We evaluate compression quality from two complementary perspectives. Direct accuracy, denoted by $D$, measures task performance using the compressed text. We send the compressed text and the question to a receiver LLM and score its answers using the dataset-specific metric defined in §\ref{sec:metrics}. Reconstruction accuracy, denoted by $R$, measures task performance after a reconstruction LLM first rebuilds the original passage from the compressed text and a receiver LLM then answers from the reconstructed text. A compressor might retain enough information for direct question answering yet lose details required by a different model for reconstruction, or vice versa.

A qualified compressor should perform well on both paths. We therefore define the fitness as:
\begin{equation}
f = \min(D, R)
\end{equation}
The $\min(D, R)$ term provides a conservative objective that focuses evolutionary pressure strictly on the weaker path. A compressor cannot achieve high fitness unless both accuracies are reasonably high, so improving the lower score is the most direct way to increase the minimum. Following evolution, we select the top-performing candidate that meets the target compression ratio and run the selected compressor on the test set.

\section{Experimental Setup}

\subsection{Datasets}
\label{sec:datasets}

We evaluate our evolved compressors across four datasets spanning medium-length passages and long documents. Specifically, \textbf{RACE}~\cite{lai-etal-2017-race} provides medium-length reading comprehension challenges from English exams. To assess long documents, we use the LongBench subset of \textbf{Qasper}~\cite{dasigi-etal-2021-dataset,bai-etal-2024-longbench}, which features queries whose answers are distributed across NLP research papers. For reasoning across multiple texts, \textbf{Multi-doc QA} concatenates supporting documents from HotpotQA~\cite{yang-etal-2018-hotpotqa} and 2WikiMultihopQA~\cite{ho-etal-2020-constructing} to enforce complex logical connections. Finally, \textbf{LongMemEval}~\cite{wu2025longmemeval} tests retention in loosely structured dialogue histories, where highly uneven information density introduces a distinct robustness challenge.


We randomly split a balanced sample from each dataset into 100 training and 200 test examples. The four datasets thus contribute equally, totaling 400 training and 800 test examples. Appendix~\ref{sec:Dataset Information} provides dataset lengths and example questions.

To assess transferability, we evaluate the evolved compressors in three out-of-distribution (OOD) settings. The OOD QA setting is constructed from LongBench~\cite{bai-etal-2024-longbench} by sampling 200 \textbf{MuSiQue} and 200 \textbf{NarrativeQA} examples. The other two settings are \textbf{long-document summarization}~\cite{cohan-etal-2018-discourse} and \textbf{20 Newsgroups} classification~\cite{lang1995newsweeder}, with 1,000 randomly sampled examples each. Complete results are in Appendix~\ref{sec:appendix-generalization}.

\subsection{Baselines}






We compare against eight hard prompt compression
baselines, covering perplexity-based,
classification-based, cross-attention-based, and syntax-based strategies.

\begin{itemize}[itemsep=1pt, topsep=1pt, parsep=1pt]
\item \textbf{Selective Context}~\cite{li-etal-2023-compressing}
removes low-self-information units scored by GPT-2~\cite{radford2019language}.
\item \textbf{LLMLingua}~\cite{jiang-etal-2023-llmlingua}
performs coarse-to-fine budget allocation with
iterative token pruning using Llama-2-7B~\cite{touvron2023llama2openfoundation}.
\item \textbf{LongLLMLingua}~\cite{jiang-etal-2024-longllmlingua}
extends LLMLingua to long contexts via contrastive
perplexity and document reordering using the same model, Llama-2-7B.
\item \textbf{LLMLingua-2}~\cite{pan-etal-2024-llmlingua}
casts compression as token classification using an
XLM-RoBERTa-large~\cite{conneau-etal-2020-unsupervised} trained on labels
distilled from GPT-4~\cite{openai2024gpt4technicalreport} to predict whether to preserve each token.
\item \textbf{Selection-p}~\cite{chung-etal-2024-selection}
introduces a small number of additional parameters
on the Llama-2-7B backbone and trains with self-supervised continual pre-training, producing a preserve-or-discard probability for each token.
\item \textbf{FrugalPrompt}~\cite{raiyan2025frugalpromptreducingcontextualoverhead}
ranks tokens by GlobEnc~\cite{modarressi-etal-2022-globenc} attention-rollout saliency over an Electra-large encoder~\cite{clark2020electrapretrainingtextencoders}.
\item \textbf{R2C}~\cite{choi-etal-2024-reading}
uses FiD decoder cross-attention to rank salient chunks
and sentences.
\item \textbf{PartPrompt}~\cite{Mao2026-px} builds a syntactic parse tree for each sentence,
computes node-level information entropy with
Llama-2-7B, organizes the local trees into a global
hierarchical tree, and prunes it recursively to
produce the compressed prompt.
\end{itemize}


\subsection{Evaluation Metrics}
\label{sec:metrics}

Direct accuracy and Reconstruction accuracy in §\ref{sec:dual-eval} are the primary metrics. For each dataset, the two paths are evaluated with the same metric.
 
 \textbf{Accuracy.} For RACE with multiple-choice QA pairs, we report exact match rate. For Qasper, Multi-doc QA, and LongMemEval with open-ended QA pairs, Direct accuracy and Reconstruction accuracy refer to LLM-as-judge accuracy, while token-level F1 serves as a reference metric for analysis. During evolutionary search, accuracy on the three open-ended datasets is based solely on the LLM-as-judge score.

\textbf{Compression ratio.} We define the compression ratio as the original
word count divided by the compressed word count.

\subsection{Implementation Details}
\label{sec:impl}

\textbf{Models.} Three open-source LLMs of comparable scale are used as both receiver LLMs and reconstruction LLMs: Qwen-2.5-7B-Instruct~\cite{qwen2025qwen25technicalreport}, LLaMA-3.1-8B-Instruct~\cite{grattafiori2024llama3herdmodels}, and Gemma-3-12B-IT~\cite{gemmateam2025gemma3technicalreport}. For each sample, one model serves as the receiver LLM and a different model serves as the reconstruction LLM; after reconstruction, a receiver LLM answers the QA prompt from the reconstructed text. GPT-5.4-mini~\cite{gpt54mini} serves as the LLM-as-judge model for the three open-ended datasets. During evolution, a three-tier ensemble comprising GPT-5.4-mini, DeepSeek-V4-Pro~\cite{dsv4}, and GPT-5.5~\cite{gpt55} proposes mutations according to the stage- and temperature-dependent schedule detailed in Appendix~\ref{sec:appendix-evolution-design}.


\textbf{Evolutionary search.} The search runs for 30 generations. Each island maintains a $6 \times 8$
MAP-Elites grid. Migration occurs every 3 generations with a migration rate of 15\%.

\begin{table*}[t]
\centering
\fontsize{8}{10}\selectfont
\setlength{\tabcolsep}{3pt}
\begin{tabular}{l cccc cccc cccc cccc}
\toprule
& \multicolumn{4}{c}{\textbf{RACE}} & \multicolumn{4}{c}{\textbf{Qasper}} & \multicolumn{4}{c}{\textbf{Multi-doc QA}} & \multicolumn{4}{c}{\textbf{LongMemEval}} \\
\cmidrule(lr){2-5} \cmidrule(lr){6-9} \cmidrule(lr){10-13} \cmidrule(lr){14-17}
Method & 1.5$\times$ & 2.5$\times$ & 4$\times$ & 6$\times$ & 1.5$\times$ & 2.5$\times$ & 4$\times$ & 6$\times$ & 1.5$\times$ & 2.5$\times$ & 4$\times$ & 6$\times$ & 1.5$\times$ & 2.5$\times$ & 4$\times$ & 6$\times$ \\
\midrule
\rowcolor{perpcolor} Selective-Ctx & \textbf{90.51} & 81.02 & 74.02 & 69.48 & \dashuline{61.50} & 49.01 & 43.00 & 24.50 & 57.07 & 56.58 & 49.54 & 45.00 & 44.04 & 37.52 & 43.52 & \underline{45.00} \\
\rowcolor{perpcolor} LLMLingua & 88.48 & \dashuline{81.98} & \textbf{78.99} & \underline{77.00} & 31.47 & 24.04 & 19.52 & 19.00 & 55.55 & 44.50 & 39.96 & 41.49 & 42.51 & 39.05 & 39.47 & 41.04 \\
\rowcolor{perpcolor} LongLLMLingua & \dashuline{89.97} & 81.48 & \underline{76.97} & \textbf{77.99} & 33.98 & 22.00 & 18.98 & 20.50 & 58.04 & 49.00 & 43.96 & 42.99 & 43.02 & 39.08 & 42.48 & 44.01 \\
\midrule
\rowcolor{clscolor} LLMLingua-2 & 88.52 & \textbf{87.00} & \dashuline{75.98} & 67.97 & 60.98 & \textbf{57.48} & \underline{50.51} & \underline{47.05} & \dashuline{60.03} & 55.03 & \underline{54.53} & \dashuline{48.56} & \dashuline{45.02} & 42.05 & 39.57 & 41.08 \\
\rowcolor{clscolor} FrugalPrompt & \underline{90.03} & 76.47 & 69.46 & 69.48 & 60.97 & 46.46 & 41.93 & \dashuline{42.97} & 58.53 & \textbf{60.05} & \dashuline{51.58} & \underline{50.09} & 37.01 & 46.54 & \underline{49.04} & \textbf{47.04} \\
\rowcolor{clscolor} R2C & 88.37 & \underline{86.33} & 75.37 & 69.03 & 59.23 & \underline{55.80} & \textbf{51.90} & \textbf{48.17} & \underline{62.17} & \underline{59.87} & \textbf{58.73} & \textbf{56.70} & 36.60 & 37.50 & 38.33 & 39.27 \\
\rowcolor{clscolor} Selection-p & 78.97 & 73.48 & 69.46 & 69.48 & 51.96 & 47.04 & 34.52 & 36.99 & 52.04 & 44.01 & 39.03 & 31.53 & \textbf{46.01} & \underline{49.03} & 44.57 & 44.08 \\
\midrule
\rowcolor{syncolor} PartPrompt & 80.00 & 78.47 & 74.54 & \dashuline{69.99} & \underline{61.99} & 49.46 & 40.49 & 36.01 & 52.55 & 47.57 & 41.53 & 37.52 & \underline{45.05} & \dashuline{48.03} & \textbf{51.03} & 42.52 \\
\midrule
\rowcolor{ourscolor} \textbf{Ours} & 86.01 & 69.52 & 66.00 & 59.48 & \textbf{62.51} & \dashuline{54.00} & \dashuline{46.48} & 39.49 & \textbf{62.51} & \dashuline{57.00} & 51.02 & 43.02 & 42.94 & \textbf{49.48} & \dashuline{48.48} & \dashuline{44.47} \\
\bottomrule
\end{tabular}
\caption{Direct accuracy (\%) across four datasets and compression ratios, averaged over three receiver LLMs. RACE uses exact match; the three open-ended datasets use LLM-as-judge accuracy as the primary metric. Table~\ref{tab:recon-results} reports the corresponding Reconstruction accuracy; full per-model results and reference token-level F1 are in Appendix~\ref{sec:appendixC}. \textbf{Bold}\,=\,best; \underline{underline}\,=\,second; \dashuline{dashed}\,=\,third.}
\label{tab:main-results}
\end{table*}

\begin{table*}[t]
\centering
\fontsize{8}{10}\selectfont
\setlength{\tabcolsep}{3pt}
\begin{tabular}{l cccc cccc cccc cccc}
\toprule
& \multicolumn{4}{c}{\textbf{RACE}} & \multicolumn{4}{c}{\textbf{Qasper}} & \multicolumn{4}{c}{\textbf{Multi-doc QA}} & \multicolumn{4}{c}{\textbf{LongMemEval}} \\
\cmidrule(lr){2-5} \cmidrule(lr){6-9} \cmidrule(lr){10-13} \cmidrule(lr){14-17}
Method & 1.5$\times$ & 2.5$\times$ & 4$\times$ & 6$\times$ & 1.5$\times$ & 2.5$\times$ & 4$\times$ & 6$\times$ & 1.5$\times$ & 2.5$\times$ & 4$\times$ & 6$\times$ & 1.5$\times$ & 2.5$\times$ & 4$\times$ & 6$\times$ \\
\midrule
\rowcolor{perpcolor} Selective-Ctx & 81.54 & \dashuline{73.03} & \textbf{70.03} & 58.49 & 35.51 & 32.99 & 33.48 & 22.49 & 37.50 & 33.52 & 32.03 & 33.49 & 45.00 & 42.53 & \dashuline{49.49} & 45.51 \\
\rowcolor{perpcolor} LLMLingua & \textbf{87.03} & 69.01 & \dashuline{66.03} & \textbf{64.04} & 26.01 & 19.52 & 14.99 & 13.50 & 28.05 & 28.04 & 30.02 & 29.49 & 44.02 & 46.03 & 46.02 & 45.55 \\
\rowcolor{perpcolor} LongLLMLingua & 84.02 & 69.50 & 64.01 & \underline{63.04} & 24.51 & 17.50 & 15.00 & 18.52 & 31.01 & 28.52 & 31.02 & 28.98 & \dashuline{48.53} & 41.49 & 43.02 & 44.57 \\
\midrule
\rowcolor{clscolor} LLMLingua-2 & \underline{85.04} & \underline{75.53} & 65.54 & 60.54 & \underline{37.00} & \underline{41.51} & \dashuline{40.49} & \dashuline{37.54} & 37.04 & \underline{40.55} & \textbf{39.03} & 35.52 & 41.02 & \textbf{54.00} & 48.53 & \dashuline{48.50} \\
\rowcolor{clscolor} FrugalPrompt & 80.51 & 64.99 & 64.02 & 57.05 & \textbf{40.02} & 38.99 & 37.98 & \underline{39.47} & \underline{39.54} & \textbf{47.53} & \underline{37.54} & \underline{40.06} & 40.03 & 45.02 & 49.04 & 47.01 \\
\rowcolor{clscolor} R2C & \dashuline{85.00} & \textbf{78.50} & \underline{67.60} & \dashuline{62.60} & 29.33 & \dashuline{40.57} & \textbf{41.03} & \textbf{43.00} & \dashuline{38.20} & 36.50 & \dashuline{37.03} & \textbf{42.53} & 43.43 & 33.30 & 39.97 & 38.23 \\
\rowcolor{clscolor} Selection-p & 69.97 & 63.01 & 61.51 & 61.99 & \textbf{40.02} & \textbf{42.49} & \underline{40.98} & 30.98 & 33.02 & 32.00 & 30.04 & 30.54 & 45.56 & 46.54 & 48.53 & \textbf{53.08} \\
\midrule
\rowcolor{syncolor} PartPrompt & 74.05 & 63.54 & 65.07 & 57.00 & \dashuline{36.01} & 36.53 & 34.97 & 28.46 & \textbf{40.56} & \dashuline{39.49} & 33.03 & \dashuline{40.03} & \underline{50.01} & \dashuline{49.50} & \textbf{50.52} & \underline{50.54} \\
\midrule
\rowcolor{ourscolor} \textbf{Ours} & 82.51 & 72.00 & 56.56 & 56.49 & 33.04 & 28.03 & 32.50 & 26.51 & 33.94 & 38.49 & 34.53 & 36.48 & \textbf{54.02} & \underline{52.00} & \underline{49.54} & 47.00 \\
\bottomrule
\end{tabular}
\caption{Reconstruction accuracy (\%) across four datasets and compression ratios, averaged over three receiver LLMs. RACE uses exact match; the three open-ended datasets use LLM-as-judge accuracy as the primary metric. Full per-model results and reference token-level F1 remain in Appendix~\ref{sec:appendixC}. \textbf{Bold}\,=\,best; \underline{underline}\,=\,second; \dashuline{dashed}\,=\,third.}
\label{tab:recon-results}
\end{table*}

\begin{figure*}[t]
  \centering
  \includegraphics[width=\textwidth]{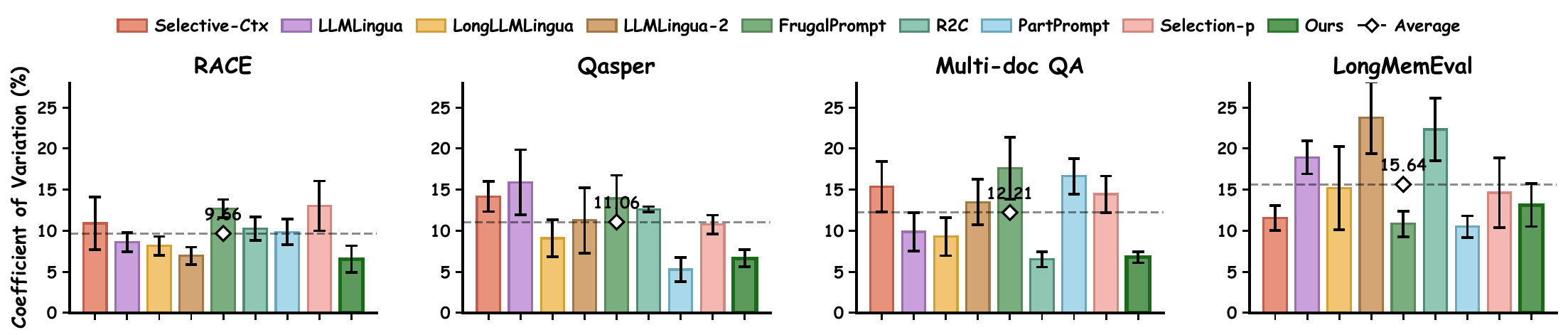}
  \caption{Cross-model stability measured by the coefficient of variation (CV, \%) of Direct accuracy across three receiver LLMs. Bars show the mean CV over four compression ratios. Lower is better.}
  \label{fig:cross-model-stability}
\end{figure*}

\textbf{Compression ratios.} We evaluate both the baselines and our evolved compressors at four target
compression ratios ($1.5\times$, $2.5\times$, $4\times$, and $6\times$). An evolved compressor does not always hit its target ratio exactly, so we associate each target with an acceptance window of
$\pm0.25$, $\pm0.5$, $\pm0.75$, and $\pm1.0$, respectively. During the search, an evolved compressor is stored in the archive only when its
compression ratio falls within this window and it passes the cascade filtering.

\section{Results and Analysis}

\subsection{Main Results}

\textbf{Linguistic rules reach a similar performance range in many lengthy-document settings.} On the three longer datasets, Table~\ref{tab:main-results} shows our compressor attaining the best Direct accuracy on 3 of the 12 dataset--ratio cells. RACE is a clear weak point: its multiple-choice passages are markedly shorter, and our method trails stronger baselines there. Because the training pool draws three quarters of its signal from longer datasets, evolution may be biased toward lengthy-text patterns. A bootstrap analysis over the original seven baselines (Appendix~\ref{sec:appendix-significance}) finds significant losses on RACE at $2.5\times$, $4\times$, and $6\times$. On the three longer datasets, the Direct LLM-as-judge comparisons are statistically inconclusive in all 12 cells, and the reference token-level F1 comparisons are statistically inconclusive in 11 of 12 cells (Tables~\ref{tab:sig-tokenf1} and~\ref{tab:sig-llmjudge}). Inconclusive outcomes establish neither a difference nor performance equivalence.

\textbf{Performance is strongest at light-to-moderate compression and degrades then.} Within Table~\ref{tab:main-results}, the compressor's rank traces a clear arc. At $1.5\times$ it ranks first on Qasper and Multi-doc QA; at $2.5\times$ it ranks first on LongMemEval and third on the other two longer datasets. Performance then tapers at $4\times$ and is weakest at $6\times$, ranking between third and fifth. A plausible explanation is that linguistic rules make categorical keep-or-drop decisions based on fixed properties such as POS tags and dependency roles, whereas LM-based scorers can rank tokens using fine-grained continuous importance values. This granularity gap matters most under the tightest budget, when only one token in six is retained. R2C is stronger on RACE and on Multi-doc QA at $2.5\times$--$6\times$, whereas ours is consistently stronger on LongMemEval. Thus, linguistic rules do not universally dominate learned cross-attention; their main advantage is lightweight deployment without LM-based scoring. The Direct and Reconstruction paths also diverge across datasets (Tables~\ref{tab:main-results} and~\ref{tab:recon-results}). Significant shortfalls fall mainly on the Direct path for RACE and on the Reconstruction path for Qasper, while Multi-doc QA shows no significant shortfall on either path and LongMemEval reveals one at $6\times$ on the Direct reference metric.

\begin{table}[t]
\centering
\small
\setlength{\tabcolsep}{3pt}
\begin{tabular}{lrrl}
\toprule
Task & Ours & Best & Baseline \\
\midrule
OOD QA & 29.00 & 29.25 & FrugalPrompt \\
Summarization & 21.39 & 22.42 & FrugalPrompt \\
Classification & 46.70 & 47.69 & PartPrompt \\
\bottomrule
\end{tabular}
\caption{Transfer without re-evolution or retuning. Scores are percentages averaged over the tested ratios: Direct judge accuracy for OOD QA ($2.5\times$--$6\times$), Direct ROUGE-L for summarization, and Direct macro-F1 for classification ($1.5\times$--$6\times$). ``Best'' is the strongest included baseline after averaging the same cells; coverage and exclusions are detailed in Appendix~\ref{sec:appendix-generalization}.}
\label{tab:generalization-summary}
\end{table}

\textbf{The evolved compressors hold the transferability beyond trained tasks.} We apply the same compressors to OOD QA (200 MuSiQue and 200 NarrativeQA examples), long-document summarization~\cite{cohan-etal-2018-discourse}, and 20 Newsgroups classification~\cite{lang1995newsweeder} (1,000 randomly sampled examples for each non-QA task). Table~\ref{tab:generalization-summary} gives a compact Direct-path summary; full Direct and Reconstruction results are in Appendix~\ref{sec:appendix-generalization}. On OOD QA, our method obtains the highest Direct judge accuracy at $2.5\times$ and $4\times$ and ranks second at $6\times$. Its mean score is within 0.25 percentage points of FrugalPrompt. On summarization and classification, it remains close to the strongest included baseline on the reported aggregate metrics.


\textbf{Linguistic compression shows low cross-model variability on two datasets.} To quantify cross-model stability, we measure the coefficient of variation (CV) of Direct accuracy. For each (method, dataset, ratio) triple, CV is defined as:
\begin{equation}
\text{CV} = \frac{\sigma}{\mu}
\end{equation}
where $\mu$ and $\sigma$ are the mean and standard deviation of accuracy over the three receiver models. We then average CV across the four compression ratios to obtain a single stability measure per method and dataset. Our method achieves the lowest CV on RACE, remains among the most stable methods on Multi-doc QA, and is competitive on Qasper and LongMemEval.

\begin{figure}[htbp]
\centering
  \includegraphics[scale=0.325]{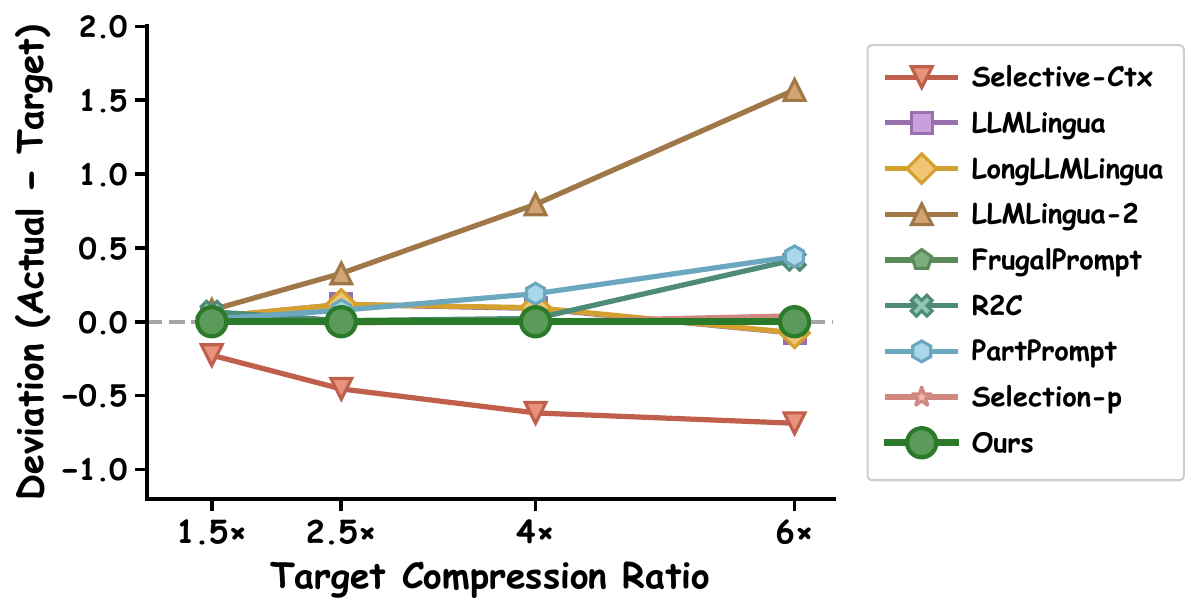}
  \caption{Deviation of the actual compression ratio from the target compression ratio (actual $-$ target), averaged over all datasets. Positive values denote over-compression.}
  \label{fig:actual_compression_ratio}
\end{figure}


\textbf{Linguistic compression keeps the compression ratio on target, whereas LM-based baselines drift.} Figure~\ref{fig:actual_compression_ratio} plots the deviation between actual and target compression ratios averaged across datasets. Our method achieves near-zero deviation at all target ratios, while several LM-based baselines exhibit drift that grows with the compression level. This controllability arises because our compressor computes an explicit word budget from the desired ratio and enforces it directly, which keeps ratio deviation small.


\textbf{The comparison involves unequal optimization budgets.} Our search uses 100 training examples from each of the same four benchmark families used at test time, whereas baselines such as LLMLingua-2 are trained once on general data rather than tuned per benchmark. The 200-example test sets are disjoint and the OOD results provide an additional check, but this asymmetry remains and is discussed in the Limitations.

\subsection{Evolutionary Analysis}
\label{evolutionary_Analysis}

\begin{figure}[t]
\centering
  \includegraphics[scale=0.80]{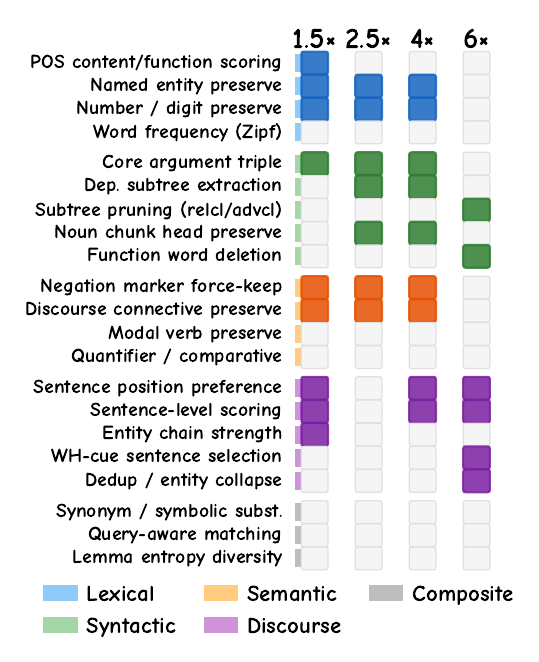}
  \caption{Survival of linguistic rules in the best evolved compressor at each compression ratio. Filled cells denote presence; rows are grouped by linguistic level. Bottom group: signals eliminated across all ratios.}
  \label{fig:signal_heatmap}
\end{figure}

\textbf{Evolution converges on a compact rule core, discarding several theoretically motivated rules.} Figure~\ref{fig:signal_heatmap} summarizes linguistic rule survival during the search at each compression ratio (details are shown in Appendix~\ref{sec:appendix-seeds}). At ratios from $1.5\times$ to $4\times$, the best compressors draw on rules from all four linguistic levels, so \textit{the best compressors found by our search consistently rely on cross-level fusion rather than any single level.} Within this range, named entity preservation, negation force-keeping, core argument triples, and discourse connective preservation persist. Each covers a distinct facet of QA-relevant information. At $6\times$, the best compressor is dominated by discourse and syntactic operations and functions as a skeleton that preserves the most critical entities while discarding finer lexical and semantic detail. Conversely, word-frequency scoring, synonym substitution, lemma entropy diversity, and query-aware matching are eliminated at every ratio. They are redundant within the final combination, not generally ineffective; Appendices~\ref{sec:appendix-qualitative} and~\ref{sec:appendix-retention-cases} provide comparisons and cases.

\textbf{The dominant unit of compression shifts from tokens to sentences as the ratio increases.} This shift is visible across the four best evolved compressors (details in Appendix~\ref{sec:best_compressor}). At $1.5\times$, the best compressor uses flat token-level scoring; at $2.5\times$, it shifts to tiered filling based on dependency subtree skeletons; at $4\times$, a two-stage cascade emerges that first selects sentences and then prunes tokens within them; and at $6\times$, the architecture becomes sentence selection followed by subtree-level deletion. Discourse-level rules track this progression, evolving from an auxiliary bonus at low ratios to the primary selection mechanism at high ratios.

\textbf{Evolutionary search generates refined mechanisms beyond any single seed.} Beyond recombining seeds, the search produces mechanisms with no direct equivalent among the initial individual seeds: dynamic entity chain scoring based on cross-sentence frequency ($1.5\times$), a hybrid truncation strategy balancing priority selection with evenly spaced document coverage ($2.5\times$), sentence-then-token composition that combines discourse and syntactic seeds into a tighter pipeline ($4\times$), and subtree-level deletion paired with a post-processing chain of entity collapse and enumeration compression ($6\times$). These innovations show that evolutionary search does not merely select predefined strategies, but can also refine and recombine them into new compositional structures through mutation.


\section{Conclusion}

Offline evolutionary search can discover linguistic rule-based compressors that require no LM forward pass at deployment. The evolved programs achieve performance within a range similar to that of recent methods in many longer-document settings, although short documents and aggressive compression remain challenging. They fuse linguistic levels and shift from token- to sentence-level operations as compression tightens, offering interpretable and lightweight prompt compression.

\section*{Limitations}

\textbf{Language and parser dependence.} This work targets English only and uses spaCy's English pipeline for tokenization, parsing, and named entity recognition. Comparable annotations from NLTK or Stanza could be mapped through an adapter, but differences in tag and dependency inventories would still require validation and a new evolutionary search for the target language. The current compressors are not directly applicable to mixed-language contexts, where tokenization and parsing errors may compound. More broadly, noisy parser outputs may change absolute performance and the rule combinations preferred by evolution.

\textbf{Input format.} Our compressors are designed for standard prose. They have not been validated on inputs with non-standard structure, such as source code, nested JSON, mathematical notation, tables, or fragmented raw data, where conventional syntactic and discourse parsing may degrade.

\textbf{Search-space coverage and distribution.} The 42 seeds do not exhaust linguistically motivated compression, and the six-island, fixed MAP-Elites configuration explores only a finite population. It may miss other rule families or converge to benchmark-specific combinations. Moreover, three long-document datasets supply three quarters of the mixed training signal, which may bias the search toward long-text patterns and contribute to weaker RACE performance. A length-balanced search pool and broader population should be tested.

\textbf{Evaluation scope and metrics.} The primary evaluation uses balanced 200-example test subsets rather than full QA benchmarks, constrained by the number of usable LongMemEval cases. The added OOD QA, summarization, and classification experiments broaden task coverage but do not establish generality across domains. For open-ended QA, conclusions may remain sensitive to the chosen judge model and prompt despite the human validation in Appendix~\ref{sec:judge-validation}. Failure to reject the null hypothesis also does not establish equivalence; non-significant comparisons are labeled statistically inconclusive.

\textbf{Receiver drift.} Our dual-path protocol uses three small LLMs of comparable scale. The evolved rule core may not transfer unchanged to future models with substantially different sizes, training data, or architectures, and re-optimization may be needed as receiver models evolve.

\textbf{Offline optimization cost and comparison asymmetry.} The final compressor runs on CPU without an LM forward pass, but discovering the four programs required about 26 hours and US\$50. This search uses 100 training samples from each of the same four benchmark families used at test time, whereas baselines such as LLMLingua-2 are trained once on general data rather than receiving the same per-benchmark optimization budget. The train and test sets are disjoint and the OOD experiments reduce, but do not remove, this asymmetry. A more sample-efficient harness such as LEVI~\cite{tanveer2026levi}, which reports comparable or better search with lower budgets, is a promising alternative backend.

\textbf{Search configuration.} We report one configuration and do not provide task-specific hyperparameter or mutator-isolation ablations. AlphaEvolve's ablations~\cite{novikov2025alphaevolvecodingagentscientific} motivate population-based evolution and model mixtures in general, but do not validate our island count, grid, migration rate, or mutator schedule. LLaMEA's $(1+1)$ strategy~\cite{vanstein2025llamea} is simpler, but its single-elite search may converge early in our multimodal space of linguistic levels and seed types; this trade-off remains empirical rather than settled.

\section*{Ethics Statement}

We do not foresee any immediate negative ethical consequences of our research.



\bibliography{custom}

\appendix

\section{Implementation Details}
All experiments use a random seed of 42. Baseline compression methods run on a single NVIDIA A100-40GB GPU; our evolved compressors run fully on CPU.

All decoding uses temperature~0 (greedy, no top-$k$ or top-$p$) except the mutator, whose temperature anneals from 1.0 (early exploration) to 0.0 (late refinement). The mutator selects from a mixture of three model tiers driven by the current temperature: \texttt{gpt-5.4-mini} as the weak tier, \texttt{deepseek-v4-pro} as the medium tier, and \texttt{gpt-5.5} as the strong tier; higher temperatures favor the weak tier for diversity and lower temperatures favor the strong tier for incremental refinement. The mutator's max-token budget is 4096 for the weak and medium tiers and 8192 for the strong.

The three receiver LLMs are queried with temperature~0 and with max\_tokens set to 64 for RACE and 256 for the open-ended datasets: Qasper, Multi-doc QA, and LongMemEval. The reconstruction LLM uses the same three-model pool with temperature~0 and max\_tokens set to 4096. Receiver LLMs and mutators are accessed through OpenRouter.

The LLM-as-judge calls use \texttt{gpt-5.4-mini} throughout, with temperature~0 and max\_tokens set to 4. The mutator and judge calls route through OpenRouter.

Per-call timeouts are 120\,s for receivers, 250\,s for the mutator's weak/medium tiers and the judge, and 500\,s for the strong tier. Retry budgets are 5 for receivers and 3 for the mutator and judge.

\subsection{Evolutionary Search Design}
\label{sec:appendix-evolution-design}

The six islands evolve independently and exchange candidates under a ring topology. Every three generations, the top-fitness 15\% of each island migrates to its two neighboring islands. Within each island, the 48-cell MAP-Elites archive is a $6\times8$ grid whose axes are code complexity and compression ratio. A cell retains only the highest-accuracy candidate in its complexity--ratio range. The grid therefore preserves behaviorally distinct programs while simultaneously exposing the ratio--accuracy trade-off.

The mutable region is the body of \texttt{compress()} delimited by \texttt{\# EVOLVE-BLOCK-START} and \texttt{\# EVOLVE-BLOCK-END}. The mutator is instructed to leave the scaffold outside these markers byte-identical. Model routing is not conditioned on whether a parent originated from a lexical, syntactic, semantic, or discourse seed. Instead, it follows generation stage through temperature annealing: early, high-temperature generations favor the weaker tier for exploration, whereas later, low-temperature generations increasingly favor the stronger tier for refinement. Finally, seeds share a common toolkit and no template constrains their composition. Cross-level fusion is thus an observed search outcome rather than a hard-coded requirement.

\subsection{Judge Protocol and Human Validation}
\label{sec:judge-validation}

For each open-ended output, GPT-5.4-mini receives the generated answer and reference answer and produces a binary correctness verdict. We call the judge three times at temperature~0 and use majority voting; the repetitions guard against rare API-level nondeterminism. The exact prompt is reported in Table~\ref{tab:llm_as_judge_prompt}.

We randomly sampled 100 judge verdicts for human validation. Three PhD-level annotators independently reviewed each verdict, and their majority vote agreed with the judge in 83\% of cases. Inter-annotator agreement was Fleiss' $\kappa=0.7617$. Annotators saw the instruction, reference answer, and generated answer. A response was marked consistent when its core facts and key information matched the reference; paraphrase, reordering, and length differences were permitted. A response was marked inconsistent if it contained a factual error, a contradiction, a critical omission, or an off-topic answer. Annotators used no external knowledge, and grammar or style did not affect the label when the core meaning was correct. These results support the judge's usefulness but do not eliminate prompt- or model-sensitivity.

\subsection{Word- and Token-based Length}
\label{sec:word-token-ratio}

All reported actual ratios use a shared output-level definition: original word count divided by compressed-output word count. This common axis applies even when a baseline internally controls a token budget with a method-specific tokenizer. Word count also matches our spaCy-based word-level operations. As a reference for token-space interpretation across the four datasets, the average token-to-word ratio is approximately 1.39 with the GPT-2 BPE tokenizer and 1.58 with the Llama-2 tokenizer.

\section{Linguistic Seeds}
\label{sec:appendix-seeds}

This appendix provides the full specification of the 42 seeds that form the initial population for evolutionary search. Each seed is a self-contained Python function \texttt{compress(text: str) -> str} encoding one linguistically motivated compression hypothesis. All seeds share a target-ratio wrapper that enforces the desired compression ratio by truncating or back-filling the output to $\lfloor n_{\text{src}} / r \rceil$ words. The seed itself determines \emph{how to rank or select} tokens, not how many to keep.

Superscripts in Tables~\ref{tab:seed-lexical}--\ref{tab:seed-composite} denote operation types: \textsuperscript{S}\,Scoring + top-k, \textsuperscript{T}\,Tiered fill, \textsuperscript{D}\,Deletion, \textsuperscript{E}\,Extraction, \textsuperscript{R}\,Substitution, \textsuperscript{L}\,Sentence selection, \textsuperscript{C}\,Cascade, \textsuperscript{P}\,Post-processing.

\begin{table}[h]
\centering
\small
\begin{tabular}{@{}c p{2.6cm} p{3.6cm}@{}}
\toprule
\# & Seed Name & Motivation \\
\midrule
1 & Content word retain\textsuperscript{S} & Keep content words, drop function words~\cite{shannon1948mathematical} \\
2 & NER-first\textsuperscript{T} & QA answers are often named entities \\
3 & Stopword deletion\textsuperscript{D} & Closed-class words carry minimal info \\
4 & Content-func.\ gap\textsuperscript{S} & Maximize content/function keep-rate gap \\
5 & Zipf rare boost\textsuperscript{S} & Rare words carry more information \\
6 & Acronym preserve\textsuperscript{S} & Acronyms are key identifiers \\
7 & Lemma dedup\textsuperscript{S} & One instance per lemma \\
8 & Synonym subst.\textsuperscript{R} & Replace with shorter synonyms \\
9 & Symbolic abbrev.\textsuperscript{R} & Replace phrases with symbols \\
10 & Canonical POS\textsuperscript{S} & POS hierarchy \\
\bottomrule
\end{tabular}
\caption{Lexical seeds (10).}
\label{tab:seed-lexical}
\end{table}

\begin{table}[h]
\centering
\small
\begin{tabular}{@{}c p{2.6cm} p{3.6cm}@{}}
\toprule
\# & Seed Name & Motivation \\
\midrule
11 & Arg.\ triple only\textsuperscript{E} & Keep (nsubj, ROOT, dobj)~\cite{10.1145/1409360.1409378} \\
12 & Verb skeleton\textsuperscript{T} & ROOT subtrees strongest at light comp. \\
13 & Dep.\ role priority\textsuperscript{S} & Core arguments above modifiers \\
14 & Relcl pruning\textsuperscript{D} & Relative clauses dispensable~\cite{filippova-strube-2008-dependency} \\
15 & Amod pruning\textsuperscript{D} & Adj.\ modifiers rarely carry answers \\
16 & Advmod pruning\textsuperscript{D} & Adv.\ modifiers highly droppable \\
17 & Conjunct pruning\textsuperscript{D} & Beyond-first conjuncts dispensable \\
18 & Appos pruning\textsuperscript{D} & Appositives restate head noun \\
19 & SVO extraction\textsuperscript{E} & Minimal propositional content \\
20 & Chunk head retain\textsuperscript{T} & Chunk head most informative \\
21 & Depth pruning\textsuperscript{D} & Deeper tokens more dispensable \\
22 & Parenthetical del.\textsuperscript{D} & Parentheticals non-essential \\
\bottomrule
\end{tabular}
\caption{Syntactic seeds (12).}
\label{tab:seed-syntactic}
\end{table}

\begin{table}[h]
\centering
\small
\begin{tabular}{@{}c p{2.6cm} p{3.6cm}@{}}
\toprule
\# & Seed Name & Motivation \\
\midrule
23 & Negation keep\textsuperscript{S} & Removal reverses polarity~\cite{horn1989negation} \\
24 & Modal preserve\textsuperscript{S} & Encodes certainty / obligation \\
25 & Quantifier preserve\textsuperscript{S} & Constrains scope~\cite{barwise-cooper-1981} \\
26 & Comparative preserve\textsuperscript{S} & Encodes relational meaning \\
27 & Reasoning anchor\textsuperscript{T} & Modal+ROOT+connective skeleton \\
28 & WH/causal boost\textsuperscript{S} & Reasoning anchors for QA \\
\bottomrule
\end{tabular}
\caption{Semantic seeds (6).}
\label{tab:seed-semantic}
\end{table}

\begin{table}[h]
\centering
\small
\begin{tabular}{@{}c p{2.6cm} p{3.6cm}@{}}
\toprule
\# & Seed Name & Motivation \\
\midrule
29 & First-sent.\ anchor\textsuperscript{T} & First-sent.\ kept at 2--5$\times$ rate \\
30 & Position decay\textsuperscript{S} & Geometric budget from first sent. \\
31 & Entity chain\textsuperscript{S} & Entities in $\geq$2 sents are anchors~\cite{halliday2014cohesion} \\
32 & TextRank\textsuperscript{L} & Centrality by word overlap \\
33 & Dedup removal\textsuperscript{L} & High-overlap sents redundant \\
34 & Causal conn.\textsuperscript{S} & Reasoning coherence~\cite{prasad-etal-2008-penn} \\
35 & Temporal chain\textsuperscript{S} & Anchors event sequences \\
36 & Doc-initial bias\textsuperscript{S} & First 20 tokens get bonus \\
37 & Entity collapse\textsuperscript{P} & Collapse repeated mentions \\
\bottomrule
\end{tabular}
\caption{Discourse seeds (9).}
\label{tab:seed-discourse}
\end{table}

\begin{table}[h]
\centering
\small
\begin{tabular}{@{}c p{2.6cm} p{3.6cm}@{}}
\toprule
\# & Seed Name & Motivation \\
\midrule
38 & Rank+dep-core\textsuperscript{C} & TextRank then dep-core extraction \\
39 & Rank+SVO\textsuperscript{C} & Sent.\ selection then SVO triples \\
40 & Entropy diversity\textsuperscript{S} & Diminishing returns for repeated lemmas \\
41 & Query-aware\textsuperscript{S} & Boost tokens matching question \\
42 & Full cascade\textsuperscript{C} & Rank $\to$ dep.\ core $\to$ SVO $\to$ symbols \\
\bottomrule
\end{tabular}
\caption{Composite pipeline seeds (5).}
\label{tab:seed-composite}
\end{table}
\subsection{Design Principles}

The 42 seeds are designed to provide the evolutionary search with a diverse initial population spanning the compression--accuracy trade-off space. Several principles guide their construction:

\textbf{Theoretical grounding.} Each seed encodes a hypothesis rooted in linguistic theory or empirical findings. Lexical seeds draw on the content--function word distinction~\cite{shannon1948mathematical,zipf2016human}; syntactic seeds on dependency grammar and sentence compression~\cite{tesniere2015elements,filippova-strube-2008-dependency}; semantic seeds on negation scope~\cite{horn1989negation} and quantifier semantics~\cite{barwise-cooper-1981}; discourse seeds on coherence theory~\cite{halliday2014cohesion} and rhetorical structure~\cite{william1988rhetorical}.

\textbf{Intentional overlap.} Several seeds share machinery (e.g., SVO extraction and entailment extraction are nearly identical; the four discourse-marker seeds all reduce to sentence-level predicates). This redundancy is intentional: it provides the evolutionary search with overlapping starting points that can be recombined through crossover and mutation.

\textbf{Level diversity.} Seeds are distributed across all four linguistic levels plus composite pipelines, ensuring that no single level dominates the initial population. This diversity is critical for the evolutionary search to discover cross-level combinations rather than converging on a single-level strategy.

\subsection{Signal Survival Across Compression Ratios}
Tables~\ref{tab:survival-lex-syn} and~\ref{tab:survival-sem-dis-comp} track which seed rules survived in the best evolved compressor at each target ratio. Rules that persist across all tested ratios represent the irreducible core of linguistic compression; those eliminated across all ratios were not beneficial for downstream QA despite their theoretical motivation.

\begin{table}[h]
\centering
\small
\setlength{\tabcolsep}{2.5pt}
\begin{tabular}{@{}c l cccc@{}}
\toprule
\# & Seed & 1.5$\times$ & 2.5$\times$ & 4$\times$ & 6$\times$ \\
\midrule
\multicolumn{6}{l}{\textit{Lexical (10)}} \\
1 & Content word retain & \checkmark & & & \\
2 & NER-first & \checkmark & \checkmark & \checkmark & \\
3 & Stopword deletion & \checkmark & & & \checkmark \\
4 & Content-func.\ gap & \checkmark & & & \\
5 & Zipf rare boost & & & & \\
6 & Acronym preserve & & \checkmark & & \\
7 & Lemma dedup & & & & \\
8 & Synonym subst. & & & & \\
9 & Symbolic abbrev. & & & & \\
10 & Canonical POS & \checkmark & & & \\
\midrule
\multicolumn{6}{l}{\textit{Syntactic (12)}} \\
11 & Arg.\ triple only & \checkmark & \checkmark & \checkmark & \\
12 & Verb skeleton & \checkmark & \checkmark & \checkmark & \\
13 & Dep.\ role priority & \checkmark & \checkmark & \checkmark & \\
14 & Relcl pruning & & & & \checkmark \\
15 & Amod pruning & & & & \\
16 & Advmod pruning & & & & \\
17 & Conjunct pruning & & & & \\
18 & Appos pruning & & & & \checkmark \\
19 & SVO extraction & & & & \\
20 & Chunk head retain & & \checkmark & \checkmark & \\
21 & Depth pruning & & & & \\
22 & Parenthetical del. & & & & \\
\bottomrule
\end{tabular}
\caption{Survival of lexical and syntactic seeds. \checkmark\,= present in the best evolved compressor.}
\label{tab:survival-lex-syn}
\end{table}

\begin{table}[h]
\centering
\small
\setlength{\tabcolsep}{2.5pt}
\begin{tabular}{@{}c l cccc@{}}
\toprule
\# & Seed & 1.5$\times$ & 2.5$\times$ & 4$\times$ & 6$\times$ \\
\midrule
\multicolumn{6}{l}{\textit{Semantic (6)}} \\
23 & Negation keep & \checkmark & \checkmark & \checkmark & \\
24 & Modal preserve & & & & \\
25 & Quantifier preserve & & & & \\
26 & Comparative preserve & & & & \\
27 & Reasoning anchor & \checkmark & \checkmark & \checkmark & \\
28 & WH/causal boost & \checkmark & \checkmark & \checkmark & \checkmark \\
\midrule
\multicolumn{6}{l}{\textit{Discourse (9)}} \\
29 & First-sent.\ anchor & \checkmark & & & \checkmark \\
30 & Position decay & \checkmark & & & \\
31 & Entity chain & \checkmark & & & \\
32 & TextRank & & & & \\
33 & Dedup removal & & & & \\
34 & Causal conn. & \checkmark & \checkmark & \checkmark & \\
35 & Temporal chain & \checkmark & \checkmark & \checkmark & \\
36 & Doc-initial bias & \checkmark & & & \checkmark \\
37 & Entity collapse & & & & \checkmark \\
\midrule
\multicolumn{6}{l}{\textit{Composite (5)}} \\
38 & Rank+dep-core & & & \checkmark & \\
39 & Rank+SVO & & & & \\
40 & Entropy diversity & & & & \\
41 & Query-aware & & & & \\
42 & Full cascade & & & & \\
\bottomrule
\end{tabular}
\caption{Survival of semantic, discourse, and composite seeds. \checkmark\,= present in the best evolved compressor.}
\label{tab:survival-sem-dis-comp}
\end{table}

\section{Dataset Information}
\label{sec:Dataset Information}

\begin{table}[h]
\centering
\small
\setlength{\tabcolsep}{6pt}
\begin{tabular}{@{}l r r@{}}
\toprule
Dataset & Train words & Test words \\
\midrule
RACE                                            & $292 \pm 131$         & $325 \pm 148$ \\
Qasper                                          & $4{,}070 \pm 2{,}273$ & $3{,}933 \pm 2{,}280$ \\
Multi-doc QA                     & $7{,}020 \pm 3{,}350$ & $6{,}582 \pm 3{,}401$ \\
LongMemEval                                     & $3{,}972 \pm 2{,}192$ & $4{,}361 \pm 2{,}459$ \\
\bottomrule
\end{tabular}
\caption{Context lengths for each dataset's training ($n=100$ per dataset) and test ($n=200$ per dataset) splits, reported as mean $\pm$ standard deviation.}
\label{tab:dataset-stats}
\end{table}

For each of the four datasets, we use $100$ training examples and $200$ test examples. Table~\ref{tab:dataset-stats} reports context length statistics (whitespace-separated word count). The four datasets span a wide range of input lengths, from short multiple-choice passages on RACE ($\sim 300$ words) to long multi-document and dialogue contexts ($\sim 4{,}000$--$7{,}000$ words). Below we also show one representative example per dataset; for long contexts the passage is truncated with ``\ldots'' for readability.

\subsection{Example: RACE}

\begin{mdframed}[backgroundcolor=promptbg, linecolor=promptbg, skipabove=5pt, skipbelow=5pt, innerleftmargin=8pt, innerrightmargin=8pt, innertopmargin=8pt, innerbottommargin=8pt]
\small
\textbf{Context} ($305$ words): Do you start your work by making a list of all you have to complete, from walking the dog and washing the clothes, to phoning a friend or checking your emails? Do you always take a list with you to the supermarket? Do you write down all the steps you need to see a big project through? If you're a natural list maker you can consider this habit as a good thing to do. \ldots A list makes it possible for us to spend too much time on unimportant items. \ldots

\textbf{Question:} The author introduces the topic to be discussed by \rule{0.4cm}{0.4pt}.

\textbf{Options:} (A) asking a series of questions; (B) presenting some simple facts; (C) stating his own point of view; (D) listing the benefits of a to-do list.

\textbf{Answer:} A
\end{mdframed}

\subsection{Example: Qasper}

\begin{mdframed}[backgroundcolor=promptbg, linecolor=promptbg, skipabove=5pt, skipbelow=5pt, innerleftmargin=8pt, innerrightmargin=8pt, innertopmargin=8pt, innerbottommargin=8pt]
\small
\textbf{Context} ($3{,}697$ words): Ancient Chinese is the writing language in ancient China. It is a treasure of Chinese culture which brings together the wisdom and ideas of the Chinese nation \ldots Compared with modern Chinese, ancient Chinese is more concise and shorter. \ldots We apply two NMT models, an RNN-based encoder--decoder and a Transformer-based model, to translate ancient Chinese into modern Chinese, and evaluate against several baselines. \ldots

\textbf{Question:} What NMT models did they compare with?

\textbf{Answer:} RNN-based NMT model, Transformer-NMT.
\end{mdframed}

\subsection{Example: Multi-doc QA}

\begin{mdframed}[backgroundcolor=promptbg, linecolor=promptbg, skipabove=5pt, skipbelow=5pt, innerleftmargin=8pt, innerrightmargin=8pt, innertopmargin=8pt, innerbottommargin=8pt]
\small
\textbf{Context} ($5{,}804$ words, multi-passage): \textit{Passage 1: Ian Barry (director).} Ian Barry is an Australian director of film and TV. \ldots \textit{Passage 2: Dana Blankstein.} Dana Blankstein-Cohen (born March 3, 1981) is the executive director of \ldots \textit{[$8$ supporting passages in total, including the one identifying the director of \emph{Scarecrow (1984)}.]} \ldots

\textbf{Question:} Where does the director of film \emph{Scarecrow (1984 Film)} work at?

\textbf{Answer:} High Courses for Scriptwriters and Film Directors.
\end{mdframed}

\subsection{Example: LongMemEval}
\begin{mdframed}[backgroundcolor=promptbg, linecolor=promptbg, skipabove=5pt, skipbelow=5pt, innerleftmargin=8pt, innerrightmargin=8pt, innertopmargin=8pt, innerbottommargin=8pt]
\small
\textbf{Context} ($4{,}414$ words, multi-turn dialogue history): \texttt{[Session 1]} \textbf{user:} I'm looking for some book recommendations. \ldots I just finished a $416$-page novel, but before that, I read ``The Power'' by Naomi Alderman in December, which had $341$ pages \ldots \textbf{assistant:} You're a reader who enjoys immersive, thought-provoking novels \ldots \textit{[In a later session, the user reports finishing ``The Nightingale'' by Kristin Hannah, $440$ pages, and asks for similar recommendations.]} \ldots

\textbf{Question:} What was the page count of the two novels I finished in January and March?

\textbf{Answer:} $856$ (i.e., $416 + 440$).
\end{mdframed}

\section{Detailed Results}
\label{sec:appendixC}

This appendix presents per-model, per-ratio results for all compression methods across the four evaluation datasets. For the three open-ended datasets, LLM-as-judge accuracy is the primary metric, while token-level F1 is reported as a reference metric.

\subsection{RACE}

Table~\ref{tab:race-detail} reports the per-model exact-match accuracy on RACE across all compression ratios. Since RACE is a multiple-choice task, LLM-as-judge evaluation is not applicable; the corresponding columns (DJ, RJ) are left blank.

\begin{table*}[t]
\centering

\resizebox{\textwidth}{!}{%
\begin{tabular}{@{}l rrrrr rrrrr rrrrr@{}}
\toprule
& \multicolumn{5}{c}{\textit{Gemma-3-12B}} & \multicolumn{5}{c}{\textit{LLaMA-3.1-8B}} & \multicolumn{5}{c}{\textit{Qwen-2.5-7B}} \\
\cmidrule(lr){2-6} \cmidrule(lr){7-11} \cmidrule(lr){12-16}
Method & $r$ & DA & RA & DJ & RJ & $r$ & DA & RA & DJ & RJ & $r$ & DA & RA & DJ & RJ \\
\midrule
\multicolumn{16}{l}{\textit{Target ratio $1.5\times$}} \\
Selective-Ctx & 1.31 & 92.42 & 89.39 & -- & -- & 1.30 & 79.10 & 74.63 & -- & -- & 1.30 & \textbf{100.00} & 80.60 & -- & -- \\
LLMLingua & 1.43 & 84.85 & \textbf{92.42} & -- & -- & 1.37 & 80.60 & \textbf{82.09} & -- & -- & 1.44 & \textbf{100.00} & 86.57 & -- & -- \\
LongLLMLingua & 1.43 & 84.85 & 87.88 & -- & -- & 1.37 & \textbf{85.07} & 79.10 & -- & -- & 1.44 & \textbf{100.00} & 85.07 & -- & -- \\
LLMLingua-2 & 1.57 & 92.42 & \textbf{92.42} & -- & -- & 1.56 & 80.60 & \textbf{82.09} & -- & -- & 1.59 & 92.54 & 80.60 & -- & -- \\
FrugalPrompt & 1.50 & \textbf{95.45} & 83.33 & -- & -- & 1.50 & 77.61 & 74.63 & -- & -- & 1.50 & 97.01 & 83.58 & -- & -- \\
R2C & 1.65 & 88.50 & 84.60 & -- & -- & 1.65 & 81.50 & 81.50 & -- & -- & 1.65 & 95.10 & \textbf{88.90} & -- & -- \\
PartPrompt & 1.49 & 80.30 & 83.33 & -- & -- & 1.57 & 67.16 & 64.18 & -- & -- & 1.58 & 92.54 & 74.63 & -- & -- \\
Selection-p & 1.49 & 72.73 & 63.64 & -- & -- & 1.50 & 79.10 & 79.10 & -- & -- & 1.49 & 85.07 & 67.16 & -- & -- \\
\textbf{Ours} & 1.50 & 87.88 & 84.85 & -- & -- & 1.50 & 77.61 & 80.60 & -- & -- & 1.50 & 92.54 & 82.09 & -- & -- \\
\textit{Rank} &  & 5 & 5 & -- & -- &  & 8 & 4 & -- & -- &  & 8 & 5 & -- & -- \\
\midrule
\multicolumn{16}{l}{\textit{Target ratio $2.5\times$}} \\
Selective-Ctx & 2.10 & 84.85 & 78.79 & -- & -- & 2.10 & 76.12 & 70.15 & -- & -- & 2.10 & 82.09 & 70.15 & -- & -- \\
LLMLingua & 2.11 & 77.27 & 71.21 & -- & -- & 2.11 & \textbf{82.09} & 71.64 & -- & -- & 2.11 & 86.57 & 64.18 & -- & -- \\
LongLLMLingua & 2.11 & 77.27 & 69.70 & -- & -- & 2.11 & 80.60 & 68.66 & -- & -- & 2.11 & 86.57 & 70.15 & -- & -- \\
LLMLingua-2 & 2.94 & \textbf{87.88} & \textbf{81.82} & -- & -- & 2.94 & \textbf{82.09} & 74.63 & -- & -- & 2.94 & 91.04 & 70.15 & -- & -- \\
FrugalPrompt & 2.50 & 69.70 & 63.64 & -- & -- & 2.50 & 70.15 & 62.69 & -- & -- & 2.50 & 89.55 & 68.66 & -- & -- \\
R2C & 2.75 & 86.20 & 78.50 & -- & -- & 2.75 & 76.50 & 76.50 & -- & -- & 2.75 & \textbf{96.30} & \textbf{80.50} & -- & -- \\
PartPrompt & 2.58 & 72.73 & 71.21 & -- & -- & 2.58 & 73.13 & 52.24 & -- & -- & 2.58 & 89.55 & 67.16 & -- & -- \\
Selection-p & 2.48 & 69.70 & 65.15 & -- & -- & 2.48 & 68.66 & 56.72 & -- & -- & 2.48 & 82.09 & 67.16 & -- & -- \\
\textbf{Ours} & 2.50 & 74.24 & 72.73 & -- & -- & 2.50 & 64.18 & \textbf{79.10} & -- & -- & 2.50 & 70.15 & 64.18 & -- & -- \\
\textit{Rank} &  & 6 & 4 & -- & -- &  & 9 & 1 & -- & -- &  & 9 & 9 & -- & -- \\
\midrule
\multicolumn{16}{l}{\textit{Target ratio $4.0\times$}} \\
Selective-Ctx & 3.48 & 77.27 & 75.76 & -- & -- & 3.48 & 65.67 & 62.69 & -- & -- & 3.48 & 79.10 & 71.64 & -- & -- \\
LLMLingua & 2.87 & 77.27 & 71.21 & -- & -- & 2.87 & 70.15 & \textbf{67.16} & -- & -- & 2.87 & \textbf{89.55} & 59.70 & -- & -- \\
LongLLMLingua & 2.87 & 71.21 & 65.15 & -- & -- & 2.87 & \textbf{73.13} & 65.67 & -- & -- & 2.87 & 86.57 & 61.19 & -- & -- \\
LLMLingua-2 & 4.95 & 72.73 & 72.73 & -- & -- & 4.95 & 71.64 & \textbf{67.16} & -- & -- & 4.95 & 83.58 & 56.72 & -- & -- \\
FrugalPrompt & 4.00 & 62.12 & 68.18 & -- & -- & 4.00 & 61.19 & 59.70 & -- & -- & 4.00 & 85.07 & 64.18 & -- & -- \\
R2C & 4.50$^{\dagger}$ & 81.20 & 61.50 & -- & -- & 4.50$^{\dagger}$ & 63.40 & 67.10 & -- & -- & 4.50$^{\dagger}$ & 81.50 & \textbf{74.20} & -- & -- \\
PartPrompt & 4.12 & \textbf{83.33} & \textbf{78.79} & -- & -- & 4.12 & 62.69 & 55.22 & -- & -- & 4.12 & 77.61 & 61.19 & -- & -- \\
Selection-p & 4.00 & 62.12 & 63.64 & -- & -- & 4.00 & 61.19 & 62.69 & -- & -- & 4.00 & 85.07 & 58.21 & -- & -- \\
\textbf{Ours} & 4.00 & 66.67 & 68.18 & -- & -- & 4.00 & 64.18 & 58.21 & -- & -- & 4.00 & 67.16 & 43.28 & -- & -- \\
\textit{Rank} &  & 7 & 6 & -- & -- &  & 5 & 8 & -- & -- &  & 9 & 9 & -- & -- \\
\midrule
\multicolumn{16}{l}{\textit{Target ratio $6.0\times$}} \\
Selective-Ctx & 5.43 & 65.15 & 56.06 & -- & -- & 5.43 & 53.73 & 62.69 & -- & -- & 5.43 & \textbf{89.55} & 56.72 & -- & -- \\
LLMLingua & 3.62 & 77.27 & \textbf{71.21} & -- & -- & 3.62 & 67.16 & \textbf{67.16} & -- & -- & 3.62 & 86.57 & 53.73 & -- & -- \\
LongLLMLingua & 3.62 & 75.76 & \textbf{71.21} & -- & -- & 3.62 & \textbf{68.66} & 65.67 & -- & -- & 3.62 & \textbf{89.55} & 52.24 & -- & -- \\
LLMLingua-2 & 7.74 & 62.12 & 68.18 & -- & -- & 7.74 & 64.18 & 58.21 & -- & -- & 7.74 & 77.61 & 55.22 & -- & -- \\
FrugalPrompt & 5.99 & 65.15 & 66.67 & -- & -- & 5.99 & 61.19 & 47.76 & -- & -- & 5.99 & 82.09 & 56.72 & -- & -- \\
R2C & 8.22$^{\ddagger}$ & \textbf{77.60} & 59.20 & -- & -- & 8.22$^{\ddagger}$ & 55.40 & 56.80 & -- & -- & 8.22$^{\ddagger}$ & 74.10 & \textbf{71.80} & -- & -- \\
PartPrompt & 6.21 & 68.18 & 57.58 & -- & -- & 6.21 & 67.16 & 59.70 & -- & -- & 6.21 & 74.63 & 53.73 & -- & -- \\
Selection-p & 6.09 & 65.15 & 59.09 & -- & -- & 6.09 & 53.73 & 62.69 & -- & -- & 6.09 & \textbf{89.55} & 64.18 & -- & -- \\
\textbf{Ours} & 6.00 & 56.06 & 54.55 & -- & -- & 6.00 & 53.73 & 55.22 & -- & -- & 6.00 & 68.66 & 59.70 & -- & -- \\
\textit{Rank} &  & 9 & 9 & -- & -- &  & 9 & 8 & -- & -- &  & 9 & 3 & -- & -- \\
\bottomrule
\end{tabular}%
}
\caption{Detailed results on RACE across compression ratios. $r$ = actual compression ratio; DA = direct accuracy (exact-match accuracy, \%); RA = reconstruction accuracy (exact-match accuracy, \%). RACE is a multiple-choice task; LLM-as-judge evaluation (DJ/RJ) is not applicable. For R2C at $4\times$ and $6\times$, $^{\dagger}$ and $^{\ddagger}$ indicate that 5 of 200 and 8 of 200 compression outputs, respectively, were empty; the reported $r$ averages only nonempty outputs. \textbf{Bold} indicates the best result in each column. The Rank row shows the ranking of our method among all methods for each metric.}
\label{tab:race-detail}
\end{table*}

\subsection{Qasper}

Table~\ref{tab:qasper-detail} reports the per-model reference token-level F1 and primary LLM-as-judge accuracy on Qasper across all compression ratios.

\begin{table*}[t]
\centering

\resizebox{\textwidth}{!}{%
\begin{tabular}{@{}l rrrrr rrrrr rrrrr@{}}
\toprule
& \multicolumn{5}{c}{\textit{Gemma-3-12B}} & \multicolumn{5}{c}{\textit{LLaMA-3.1-8B}} & \multicolumn{5}{c}{\textit{Qwen-2.5-7B}} \\
\cmidrule(lr){2-6} \cmidrule(lr){7-11} \cmidrule(lr){12-16}
Method & $r$ & DA & RA & DJ & RJ & $r$ & DA & RA & DJ & RJ & $r$ & DA & RA & DJ & RJ \\
\midrule
\multicolumn{16}{l}{\textit{Target ratio $1.5\times$}} \\
Selective-Ctx & 1.26 & 35.40 & 23.59 & 62.12 & 37.88 & 1.26 & 45.53 & 29.57 & 71.64 & 31.34 & 1.26 & 30.65 & 17.34 & 50.75 & 37.31 \\
LLMLingua & 1.82 & 21.54 & 10.92 & 25.76 & 27.27 & 1.89 & 27.57 & 20.93 & 41.79 & 23.88 & 1.89 & 17.23 & 11.36 & 26.87 & 26.87 \\
LongLLMLingua & 1.82 & 20.35 & 12.64 & 30.30 & 27.27 & 1.89 & 28.16 & 24.70 & 38.81 & 22.39 & 1.89 & 17.68 & 11.10 & 32.84 & 23.88 \\
LLMLingua-2 & 1.58 & 33.32 & 20.60 & 56.06 & 37.88 & 1.58 & 42.50 & 28.16 & \textbf{74.63} & 34.33 & 1.58 & 30.29 & 21.93 & 52.24 & 38.81 \\
FrugalPrompt & 1.50 & 31.74 & 20.78 & 54.55 & \textbf{43.94} & 1.50 & 46.51 & \textbf{30.69} & 67.16 & 34.33 & 1.50 & \textbf{35.18} & 20.29 & 61.19 & \textbf{41.79} \\
R2C & 1.51 & \textbf{35.70} & \textbf{28.30} & 64.40 & 41.00 & 1.51 & \textbf{47.10} & 18.10 & 63.90 & 23.50 & 1.51 & 31.40 & \textbf{24.70} & 49.40 & 23.50 \\
PartPrompt & 1.50 & 29.01 & 24.26 & 59.09 & 37.88 & 1.49 & 44.16 & 29.77 & 61.19 & 38.81 & 1.50 & 33.14 & 15.27 & \textbf{65.67} & 31.34 \\
Selection-p & 1.51 & 32.57 & 24.48 & 43.94 & \textbf{43.94} & 1.49 & 39.25 & 28.60 & 53.73 & \textbf{44.78} & 1.51 & 28.84 & 18.45 & 58.21 & 31.34 \\
\textbf{Ours} & 1.50 & 35.00 & 25.20 & \textbf{65.15} & 40.91 & 1.50 & 44.40 & 18.90 & 64.18 & 32.84 & 1.50 & 29.70 & 13.20 & 58.21 & 25.37 \\
\textit{Rank} &  & 3 & 2 & 1 & 4 &  & 4 & 8 & 4 & 5 &  & 6 & 7 & 4 & 7 \\
\midrule
\multicolumn{16}{l}{\textit{Target ratio $2.5\times$}} \\
Selective-Ctx & 2.02 & 26.78 & 16.71 & 51.52 & 30.30 & 2.02 & 38.29 & 26.31 & 52.24 & 32.84 & 2.02 & 29.77 & 17.46 & 43.28 & 35.82 \\
LLMLingua & 3.60 & 19.51 & 10.91 & 31.82 & 22.73 & 3.60 & 19.51 & 20.18 & 19.40 & 20.90 & 3.60 & 10.37 & 6.24 & 20.90 & 14.93 \\
LongLLMLingua & 3.60 & 17.40 & 8.87 & 22.73 & 18.18 & 3.60 & 21.43 & 19.81 & 20.90 & 19.40 & 3.60 & 10.33 & 9.18 & 22.39 & 14.93 \\
LLMLingua-2 & 2.78 & \textbf{35.09} & 22.93 & 53.03 & 43.94 & 2.78 & 42.52 & 31.86 & 59.70 & 41.79 & 2.78 & 28.44 & 22.24 & \textbf{59.70} & 38.81 \\
FrugalPrompt & 2.50 & 28.08 & 26.75 & 37.88 & 37.88 & 2.50 & 39.55 & \textbf{35.20} & 55.22 & \textbf{52.24} & 2.50 & 28.21 & 17.99 & 46.27 & 26.87 \\
R2C & 2.42 & 34.50 & \textbf{28.70} & 53.90 & \textbf{50.30} & 2.42 & \textbf{44.30} & 23.90 & \textbf{65.40} & 27.00 & 2.42 & 31.00 & \textbf{30.40} & 48.10 & 44.40 \\
PartPrompt & 2.58 & 23.50 & 26.31 & 42.42 & 42.42 & 2.58 & 34.53 & 26.46 & 52.24 & 31.34 & 2.58 & 30.85 & 22.90 & 53.73 & 35.82 \\
Selection-p & 2.51 & 32.70 & 23.39 & \textbf{54.55} & 40.91 & 2.51 & 33.45 & 28.02 & 40.30 & 38.81 & 2.51 & 26.88 & 19.92 & 46.27 & \textbf{47.76} \\
\textbf{Ours} & 2.50 & 30.20 & 17.40 & \textbf{54.55} & 33.33 & 2.50 & 34.90 & 25.40 & 50.75 & 28.36 & 2.50 & \textbf{39.60} & 14.80 & 56.72 & 22.39 \\
\textit{Rank} &  & 4 & 6 & 2 & 6 &  & 5 & 6 & 6 & 6 &  & 1 & 7 & 2 & 7 \\
\midrule
\multicolumn{16}{l}{\textit{Target ratio $4.0\times$}} \\
Selective-Ctx & 3.35 & 27.09 & 20.80 & 42.42 & 28.79 & 3.35 & 32.68 & 23.66 & 52.24 & 35.82 & 3.35 & 24.56 & 22.43 & 34.33 & 35.82 \\
LLMLingua & 5.61 & 21.36 & 8.88 & 22.73 & 13.64 & 5.61 & 17.72 & 17.08 & 16.42 & 13.43 & 5.61 & 9.70 & 11.67 & 19.40 & 17.91 \\
LongLLMLingua & 5.63 & 16.70 & 10.85 & 15.15 & 15.15 & 5.63 & 15.82 & 14.89 & 19.40 & 17.91 & 5.63 & 10.28 & 10.04 & 22.39 & 11.94 \\
LLMLingua-2 & 4.71 & 29.06 & 21.08 & 51.52 & 39.39 & 4.71 & 35.80 & 30.25 & 49.25 & 38.81 & 4.71 & 26.55 & 22.59 & \textbf{50.75} & 43.28 \\
FrugalPrompt & 4.00 & 21.70 & 26.39 & 28.79 & 34.85 & 4.00 & 37.74 & \textbf{35.47} & 46.27 & \textbf{46.27} & 4.00 & 27.68 & 20.07 & \textbf{50.75} & 32.84 \\
R2C & 3.88 & \textbf{32.40} & \textbf{30.20} & \textbf{52.10} & \textbf{49.10} & 3.88 & \textbf{41.20} & 27.00 & \textbf{60.40} & 33.30 & 3.88 & 31.40 & \textbf{29.80} & 43.20 & 40.70 \\
PartPrompt & 4.16 & 26.16 & 25.26 & 39.39 & 28.79 & 4.16 & 31.41 & 24.27 & 40.30 & 34.33 & 4.16 & 28.64 & 26.17 & 41.79 & 41.79 \\
Selection-p & 4.01 & 32.19 & 28.15 & 37.88 & 37.88 & 4.01 & 28.26 & 26.92 & 32.84 & 38.81 & 4.01 & 22.46 & 22.16 & 32.84 & \textbf{46.27} \\
\textbf{Ours} & 4.00 & 24.60 & 22.00 & 42.42 & 31.82 & 4.00 & 27.60 & 23.80 & 46.27 & 34.33 & 4.00 & \textbf{36.40} & 20.20 & \textbf{50.75} & 31.34 \\
\textit{Rank} &  & 6 & 5 & 4 & 5 &  & 7 & 6 & 5 & 6 &  & 1 & 6 & 3 & 7 \\
\midrule
\multicolumn{16}{l}{\textit{Target ratio $6.0\times$}} \\
Selective-Ctx & 5.38 & 20.48 & 13.44 & 24.24 & 21.21 & 5.38 & 17.41 & 16.73 & 19.40 & 14.93 & 5.38 & 20.06 & 19.83 & 29.85 & 31.34 \\
LLMLingua & 7.93 & 18.41 & 10.75 & 19.70 & 13.64 & 7.93 & 18.08 & 14.25 & 17.91 & 14.93 & 7.93 & 12.03 & 7.99 & 19.40 & 11.94 \\
LongLLMLingua & 7.92 & 17.97 & 14.99 & 19.70 & 22.73 & 7.92 & 17.51 & 15.37 & 19.40 & 14.93 & 7.92 & 13.43 & 7.65 & 22.39 & 17.91 \\
LLMLingua-2 & 7.47 & \textbf{35.91} & 24.80 & \textbf{56.06} & 45.45 & 7.47 & 32.87 & 28.81 & 32.84 & 34.33 & 7.47 & 28.45 & 21.84 & \textbf{52.24} & 32.84 \\
FrugalPrompt & 6.00 & 25.78 & 25.21 & 37.88 & 33.33 & 6.00 & 36.64 & \textbf{38.91} & 43.28 & \textbf{46.27} & 6.00 & 28.14 & 18.63 & 47.76 & 38.81 \\
R2C & 5.89 & 33.90 & \textbf{35.00} & 50.10 & \textbf{47.70} & 5.89 & \textbf{42.80} & 28.80 & \textbf{54.20} & 38.20 & 5.89 & 28.90 & \textbf{26.30} & 40.20 & \textbf{43.10} \\
PartPrompt & 6.35 & 27.34 & 20.37 & 37.88 & 21.21 & 6.35 & 27.11 & 24.52 & 34.33 & 25.37 & 6.35 & 21.39 & 22.95 & 35.82 & 38.81 \\
Selection-p & 6.05 & 19.07 & 15.95 & 34.85 & 27.27 & 6.05 & 26.67 & 20.65 & 32.84 & 32.84 & 6.05 & 22.46 & 18.91 & 43.28 & 32.84 \\
\textbf{Ours} & 6.00 & 25.10 & 16.50 & 37.88 & 28.79 & 6.00 & 26.10 & 22.80 & 44.78 & 23.88 & 6.00 & \textbf{29.70} & 20.60 & 35.82 & 26.87 \\
\textit{Rank} &  & 5 & 5 & 5 & 4 &  & 6 & 5 & 2 & 6 &  & 1 & 4 & 6 & 7 \\
\bottomrule
\end{tabular}%
}
\caption{Detailed results on Qasper across compression ratios. $r$ = actual compression ratio; DA = reference direct token-level F1 (\%); RA = reference reconstruction token-level F1 (\%); DJ = direct LLM-as-judge accuracy (\%); RJ = reconstruction LLM-as-judge accuracy (\%). \textbf{Bold} indicates the best result in each column. The Rank row shows the ranking of our method among all methods for each metric.}
\label{tab:qasper-detail}
\end{table*}

\subsection{Multi-doc QA}

Table~\ref{tab:hotpotqa2wikimqa-detail} reports the per-model reference token-level F1 and primary LLM-as-judge accuracy on Multi-doc QA across all compression ratios.

\begin{table*}[t]
\centering

\resizebox{\textwidth}{!}{%
\begin{tabular}{@{}l rrrrr rrrrr rrrrr@{}}
\toprule
& \multicolumn{5}{c}{\textit{Gemma-3-12B}} & \multicolumn{5}{c}{\textit{LLaMA-3.1-8B}} & \multicolumn{5}{c}{\textit{Qwen-2.5-7B}} \\
\cmidrule(lr){2-6} \cmidrule(lr){7-11} \cmidrule(lr){12-16}
Method & $r$ & DA & RA & DJ & RJ & $r$ & DA & RA & DJ & RJ & $r$ & DA & RA & DJ & RJ \\
\midrule
\multicolumn{16}{l}{\textit{Target ratio $1.5\times$}} \\
Selective-Ctx & 1.26 & \textbf{54.87} & 34.73 & \textbf{71.21} & 37.88 & 1.26 & 30.51 & 34.85 & 49.25 & \textbf{38.81} & 1.26 & 47.85 & 29.07 & 50.75 & 35.82 \\
LLMLingua & 1.44 & 53.91 & 34.11 & 65.15 & 37.88 & 1.42 & 46.36 & 19.93 & 55.22 & 20.90 & 1.42 & 44.88 & 29.93 & 46.27 & 25.37 \\
LongLLMLingua & 1.45 & 54.37 & 32.96 & 65.15 & 33.33 & 1.42 & 44.16 & 23.39 & 61.19 & 31.34 & 1.42 & 44.44 & 30.49 & 47.76 & 28.36 \\
LLMLingua-2 & 1.59 & 54.39 & 41.81 & 65.15 & 45.45 & 1.59 & 44.10 & 30.26 & 59.70 & 35.82 & 1.60 & 54.61 & 30.87 & 55.22 & 29.85 \\
FrugalPrompt & 1.50 & 49.58 & 45.64 & 65.15 & 48.48 & 1.50 & 44.81 & \textbf{36.13} & 58.21 & \textbf{38.81} & 1.50 & 49.21 & 26.36 & 52.24 & 31.34 \\
R2C & 1.55 & 51.00 & 30.30 & 56.40 & 26.90 & 1.55 & 43.50 & 30.20 & 67.10 & 38.30 & 1.55 & \textbf{59.10} & \textbf{46.90} & \textbf{63.00} & \textbf{49.40} \\
PartPrompt & 1.52 & 47.69 & \textbf{46.57} & 62.12 & \textbf{53.03} & 1.53 & 36.19 & 27.10 & 49.25 & 32.84 & 1.53 & 47.12 & 34.28 & 46.27 & 35.82 \\
Selection-p & 1.51 & 46.31 & 35.74 & 60.61 & 37.88 & 1.50 & 41.03 & 32.04 & 53.73 & 35.82 & 1.52 & 43.64 & 26.49 & 41.79 & 25.37 \\
\textbf{Ours} & 1.50 & 48.10 & 22.00 & 65.15 & 22.73 & 1.50 & \textbf{47.20} & 29.50 & \textbf{67.16} & 32.84 & 1.50 & 47.50 & 41.60 & 55.22 & 46.27 \\
\textit{Rank} &  & 7 & 9 & 6 & 9 &  & 1 & 6 & 1 & 7 &  & 5 & 2 & 3 & 2 \\
\midrule
\multicolumn{16}{l}{\textit{Target ratio $2.5\times$}} \\
Selective-Ctx & 2.03 & 53.01 & 33.31 & \textbf{72.73} & 37.88 & 2.03 & 24.88 & 23.91 & 41.79 & 31.34 & 2.03 & 47.67 & 24.68 & 55.22 & 31.34 \\
LLMLingua & 2.40 & 35.86 & 33.55 & 43.94 & 36.36 & 2.40 & 28.82 & 21.92 & 46.27 & 28.36 & 2.40 & 38.39 & 22.48 & 43.28 & 19.40 \\
LongLLMLingua & 2.40 & 40.35 & 30.61 & 50.00 & 33.33 & 2.40 & 27.08 & 24.00 & 50.75 & 28.36 & 2.40 & 42.81 & 25.16 & 46.27 & 23.88 \\
LLMLingua-2 & 2.86 & 43.46 & 39.22 & 60.61 & 50.00 & 2.86 & 38.83 & 30.22 & 56.72 & 35.82 & 2.86 & 43.85 & 32.79 & 47.76 & 35.82 \\
FrugalPrompt & 2.50 & \textbf{55.90} & \textbf{47.42} & 69.70 & \textbf{53.03} & 2.50 & \textbf{43.23} & \textbf{42.92} & 55.22 & \textbf{46.27} & 2.50 & \textbf{55.05} & 35.62 & 55.22 & 43.28 \\
R2C & 2.43 & 48.90 & 30.90 & 57.90 & 28.20 & 2.43 & 43.10 & 32.80 & \textbf{66.10} & 37.00 & 2.43 & 54.10 & \textbf{44.20} & 55.60 & \textbf{44.30} \\
PartPrompt & 2.63 & 50.24 & 39.14 & 62.12 & 37.88 & 2.63 & 27.12 & 31.48 & 41.79 & 37.31 & 2.63 & 41.75 & 37.37 & 38.81 & 43.28 \\
Selection-p & 2.51 & 35.00 & 25.86 & 45.45 & 31.82 & 2.51 & 27.84 & 34.41 & 38.81 & 37.31 & 2.51 & 47.93 & 25.92 & 47.76 & 26.87 \\
\textbf{Ours} & 2.50 & 46.60 & 30.60 & 56.06 & 36.36 & 2.50 & 38.90 & 35.40 & 53.73 & 43.28 & 2.50 & 51.50 & 34.60 & \textbf{61.19} & 35.82 \\
\textit{Rank} &  & 5 & 8 & 6 & 6 &  & 3 & 2 & 4 & 2 &  & 3 & 4 & 1 & 5 \\
\midrule
\multicolumn{16}{l}{\textit{Target ratio $4.0\times$}} \\
Selective-Ctx & 3.42 & 43.01 & 33.96 & 57.58 & 37.88 & 3.42 & 28.48 & 20.36 & 38.81 & 25.37 & 3.42 & 49.61 & 27.22 & 52.24 & 32.84 \\
LLMLingua & 3.98 & 30.38 & 30.40 & 31.82 & 33.33 & 3.98 & 23.92 & 22.94 & 43.28 & 26.87 & 3.98 & 40.40 & 29.36 & 44.78 & 29.85 \\
LongLLMLingua & 3.98 & 29.28 & 35.23 & 36.36 & 34.85 & 3.98 & 25.54 & 20.56 & 43.28 & 25.37 & 3.98 & 47.07 & 34.89 & 52.24 & 32.84 \\
LLMLingua-2 & 4.94 & 43.97 & 40.77 & 60.61 & \textbf{45.45} & 4.94 & \textbf{41.59} & 29.71 & \textbf{61.19} & 35.82 & 4.94 & 38.99 & 29.43 & 41.79 & 35.82 \\
FrugalPrompt & 4.00 & \textbf{56.36} & \textbf{45.44} & \textbf{68.18} & \textbf{45.45} & 4.00 & 33.21 & \textbf{32.19} & 38.81 & 34.33 & 4.00 & 44.39 & 27.35 & 47.76 & 32.84 \\
R2C & 3.89 & 50.30 & 32.20 & 61.40 & 35.90 & 3.89 & 40.90 & 23.10 & 58.00 & \textbf{39.50} & 3.89 & \textbf{50.50} & \textbf{41.60} & \textbf{56.80} & 35.70 \\
PartPrompt & 4.38 & 42.45 & 41.24 & 48.48 & 39.39 & 4.38 & 20.67 & 19.09 & 29.85 & 22.39 & 4.38 & 43.80 & 32.45 & 46.27 & \textbf{37.31} \\
Selection-p & 4.03 & 42.47 & 36.01 & 45.45 & 37.88 & 4.03 & 30.63 & 22.96 & 38.81 & 26.87 & 4.03 & 33.38 & 26.27 & 32.84 & 25.37 \\
\textbf{Ours} & 4.00 & 45.30 & 36.00 & 54.55 & 40.91 & 4.00 & 30.20 & 22.20 & 47.76 & 28.36 & 4.00 & 41.60 & 27.80 & 50.75 & 34.33 \\
\textit{Rank} &  & 3 & 5 & 5 & 3 &  & 5 & 6 & 3 & 4 &  & 6 & 6 & 4 & 4 \\
\midrule
\multicolumn{16}{l}{\textit{Target ratio $6.0\times$}} \\
Selective-Ctx & 5.41 & 44.02 & 32.18 & 45.45 & 31.82 & 5.41 & 29.47 & 24.07 & 41.79 & 29.85 & 5.41 & 42.92 & 32.29 & 47.76 & 38.81 \\
LLMLingua & 6.11 & 27.78 & 27.91 & 39.39 & 27.27 & 6.11 & 31.49 & 28.85 & 46.27 & 35.82 & 6.11 & 36.55 & 27.52 & 38.81 & 25.37 \\
LongLLMLingua & 6.13 & 27.93 & 26.81 & 40.91 & 24.24 & 6.13 & 34.07 & 30.21 & 46.27 & 37.31 & 6.13 & 37.30 & 27.29 & 41.79 & 25.37 \\
LLMLingua-2 & 7.92 & 45.96 & 34.52 & 60.61 & 39.39 & 7.92 & \textbf{39.85} & 34.69 & 49.25 & 38.81 & 7.92 & 32.63 & 28.88 & 35.82 & 28.36 \\
FrugalPrompt & 6.00 & \textbf{58.67} & \textbf{47.85} & \textbf{68.18} & \textbf{51.52} & 6.00 & 36.30 & 28.51 & 37.31 & 34.33 & 6.00 & 42.69 & 28.31 & 44.78 & 34.33 \\
R2C & 5.86 & 44.00 & 30.60 & 60.50 & 37.20 & 5.86 & 34.90 & 40.10 & \textbf{50.30} & 47.40 & 5.86 & \textbf{49.10} & \textbf{41.80} & \textbf{59.30} & \textbf{43.00} \\
PartPrompt & 6.82 & 41.86 & 39.11 & 40.91 & 45.45 & 6.82 & 23.65 & 32.70 & 31.34 & 38.81 & 6.82 & 41.78 & 33.18 & 40.30 & 35.82 \\
Selection-p & 6.06 & 34.34 & 34.88 & 37.88 & 37.88 & 6.06 & 26.06 & 26.69 & 22.39 & 28.36 & 6.06 & 34.02 & 31.07 & 34.33 & 25.37 \\
\textbf{Ours} & 6.00 & 36.30 & 25.20 & 46.97 & 31.82 & 6.00 & 30.80 & \textbf{40.30} & 38.81 & \textbf{47.76} & 6.00 & 43.00 & 25.40 & 43.28 & 29.85 \\
\textit{Rank} &  & 6 & 9 & 4 & 7 &  & 6 & 1 & 6 & 1 &  & 2 & 9 & 4 & 5 \\
\bottomrule
\end{tabular}%
}
\caption{Detailed results on Multi-doc QA across compression ratios. $r$ = actual compression ratio; DA = reference direct token-level F1 (\%); RA = reference reconstruction token-level F1 (\%); DJ = direct LLM-as-judge accuracy (\%); RJ = reconstruction LLM-as-judge accuracy (\%). \textbf{Bold} indicates the best result in each column. The Rank row shows the ranking of our method among all methods for each metric.}
\label{tab:hotpotqa2wikimqa-detail}
\end{table*}

\subsection{LongMemEval}

Table~\ref{tab:longmemeval-detail} reports the per-model reference token-level F1 and primary LLM-as-judge accuracy on LongMemEval across all compression ratios.

\begin{table*}[t]
\centering

\resizebox{\textwidth}{!}{%
\begin{tabular}{@{}l rrrrr rrrrr rrrrr@{}}
\toprule
& \multicolumn{5}{c}{\textit{Gemma-3-12B}} & \multicolumn{5}{c}{\textit{LLaMA-3.1-8B}} & \multicolumn{5}{c}{\textit{Qwen-2.5-7B}} \\
\cmidrule(lr){2-6} \cmidrule(lr){7-11} \cmidrule(lr){12-16}
Method & $r$ & DA & RA & DJ & RJ & $r$ & DA & RA & DJ & RJ & $r$ & DA & RA & DJ & RJ \\
\midrule
\multicolumn{16}{l}{\textit{Target ratio $1.5\times$}} \\
Selective-Ctx & 1.26 & 2.35 & 2.37 & 53.03 & 45.45 & 1.27 & 1.82 & 1.21 & 43.28 & 43.28 & 1.28 & 1.50 & \textbf{2.72} & 35.82 & 46.27 \\
LLMLingua & 1.41 & \textbf{6.01} & 2.87 & 45.45 & 48.48 & 1.43 & 0.84 & 2.25 & \textbf{50.75} & 43.28 & 1.42 & 2.21 & 1.37 & 31.34 & 40.30 \\
LongLLMLingua & 1.41 & \textbf{6.01} & 2.90 & 46.97 & 54.55 & 1.43 & 0.81 & 2.22 & 49.25 & 52.24 & 1.42 & 1.50 & 0.94 & 32.84 & 38.81 \\
LLMLingua-2 & 1.57 & 2.35 & 2.40 & 50.00 & 45.45 & 1.57 & 0.88 & 0.00 & 46.27 & 43.28 & 1.58 & 0.70 & 0.39 & 38.81 & 34.33 \\
FrugalPrompt & 1.50 & 4.24 & 2.49 & 39.39 & 45.45 & 1.50 & 0.20 & 1.03 & 34.33 & 32.84 & 1.50 & 4.06 & 2.15 & 37.31 & 41.79 \\
R2C & 1.56 & 2.10 & 1.80 & 33.30 & 51.30 & 1.56 & 1.30 & \textbf{2.40} & 50.60 & 46.90 & 1.56 & 2.70 & 0.40 & 25.90 & 32.10 \\
PartPrompt & 1.50 & 5.66 & 3.24 & \textbf{54.55} & 51.52 & 1.50 & 2.54 & 1.07 & 40.30 & \textbf{53.73} & 1.51 & \textbf{5.74} & 1.25 & 40.30 & 44.78 \\
Selection-p & 1.50 & 0.55 & 2.96 & 48.48 & \textbf{57.58} & 1.50 & 0.00 & \textbf{2.40} & 46.27 & 38.81 & 1.50 & 2.61 & 1.02 & 43.28 & 40.30 \\
\textbf{Ours} & 1.50 & 2.30 & \textbf{3.60} & 30.30 & \textbf{57.58} & 1.50 & \textbf{3.50} & 0.20 & \textbf{50.75} & \textbf{53.73} & 1.50 & 2.40 & 0.80 & \textbf{47.76} & \textbf{50.75} \\
\textit{Rank} &  & 7 & 1 & 9 & 2 &  & 1 & 8 & 2 & 2 &  & 5 & 7 & 1 & 1 \\
\midrule
\multicolumn{16}{l}{\textit{Target ratio $2.5\times$}} \\
Selective-Ctx & 2.03 & 2.26 & 2.25 & 42.42 & 48.48 & 2.03 & 0.27 & 0.45 & 35.82 & 31.34 & 2.03 & 0.82 & 1.11 & 34.33 & 47.76 \\
LLMLingua & 2.36 & 4.91 & 2.89 & 48.48 & 51.52 & 2.36 & 0.56 & 1.63 & 43.28 & 46.27 & 2.36 & 1.92 & 1.91 & 25.37 & 40.30 \\
LongLLMLingua & 2.36 & 4.91 & 2.58 & \textbf{54.55} & 39.39 & 2.36 & 0.43 & 0.39 & 37.31 & 46.27 & 2.36 & 2.92 & 2.24 & 25.37 & 38.81 \\
LLMLingua-2 & 2.73 & 1.71 & \textbf{3.10} & 51.52 & 53.03 & 2.73 & 0.65 & 1.89 & 44.78 & \textbf{52.24} & 2.73 & 1.69 & \textbf{2.83} & 29.85 & 56.72 \\
FrugalPrompt & 2.50 & 4.24 & 2.52 & \textbf{54.55} & 48.48 & 2.50 & 1.81 & 0.55 & 38.81 & 32.84 & 2.50 & 3.66 & 1.04 & 46.27 & 53.73 \\
R2C & 2.43 & 3.50 & 2.40 & 39.70 & 29.50 & 2.43 & 1.00 & \textbf{2.60} & 50.60 & 35.80 & 2.43 & \textbf{4.50} & 1.60 & 22.20 & 34.60 \\
PartPrompt & 2.52 & \textbf{5.56} & 2.90 & 53.03 & 50.00 & 2.52 & \textbf{2.69} & 0.59 & 41.79 & 46.27 & 2.52 & 3.38 & 1.83 & \textbf{49.25} & 52.24 \\
Selection-p & 2.47 & 1.71 & 2.87 & \textbf{54.55} & \textbf{54.55} & 2.47 & 2.26 & 1.73 & 46.27 & 38.81 & 2.47 & 2.61 & 0.89 & 46.27 & 46.27 \\
\textbf{Ours} & 2.50 & 0.50 & 2.30 & 45.45 & 51.52 & 2.50 & 2.30 & 2.20 & \textbf{55.22} & 46.27 & 2.50 & 1.30 & 2.10 & 47.76 & \textbf{58.21} \\
\textit{Rank} &  & 9 & 8 & 7 & 4 &  & 2 & 2 & 1 & 5 &  & 8 & 3 & 2 & 1 \\
\midrule
\multicolumn{16}{l}{\textit{Target ratio $4.0\times$}} \\
Selective-Ctx & 3.28 & 2.30 & 2.68 & 48.48 & 46.97 & 3.28 & 0.10 & 1.55 & 46.27 & 47.76 & 3.28 & 0.00 & 1.80 & 35.82 & \textbf{53.73} \\
LLMLingua & 3.90 & 2.35 & \textbf{3.00} & 33.33 & 50.00 & 3.90 & 1.69 & 0.67 & 37.31 & 50.75 & 3.90 & 2.32 & 1.51 & 47.76 & 37.31 \\
LongLLMLingua & 3.90 & 2.35 & 2.19 & 37.88 & 46.97 & 3.90 & 1.77 & 1.47 & 41.79 & 50.75 & 3.90 & 2.32 & 1.51 & 47.76 & 31.34 \\
LLMLingua-2 & 4.58 & 2.58 & 2.72 & 53.03 & 54.55 & 4.58 & 0.85 & \textbf{2.67} & 43.28 & 41.79 & 4.58 & 0.00 & \textbf{2.75} & 22.39 & 49.25 \\
FrugalPrompt & 4.00 & 4.24 & 2.58 & \textbf{57.58} & \textbf{57.58} & 4.00 & 1.42 & 0.80 & 46.27 & 43.28 & 4.00 & 2.52 & 0.89 & 43.28 & 46.27 \\
R2C & 3.82 & 4.50 & 2.20 & 41.00 & 37.20 & 3.82 & 0.50 & 2.60 & 44.40 & 37.00 & 3.82 & 1.80 & 1.10 & 29.60 & 45.70 \\
PartPrompt & 4.10 & \textbf{5.87} & 2.94 & \textbf{57.58} & 54.55 & 4.10 & \textbf{2.63} & 1.38 & \textbf{49.25} & 44.78 & 4.10 & \textbf{5.13} & 1.46 & 46.27 & 52.24 \\
Selection-p & 3.96 & 4.24 & 2.72 & \textbf{57.58} & 54.55 & 3.96 & 1.76 & 1.73 & 34.33 & 38.81 & 3.96 & 2.53 & 1.11 & 41.79 & 52.24 \\
\textbf{Ours} & 4.00 & 2.20 & 1.40 & 43.94 & \textbf{57.58} & 4.00 & 0.80 & 1.20 & 47.76 & \textbf{53.73} & 4.00 & 1.90 & 0.30 & \textbf{53.73} & 37.31 \\
\textit{Rank} &  & 9 & 9 & 6 & 2 &  & 7 & 7 & 2 & 1 &  & 6 & 9 & 1 & 8 \\
\midrule
\multicolumn{16}{l}{\textit{Target ratio $6.0\times$}} \\
Selective-Ctx & 5.03 & 3.52 & 3.10 & 45.45 & 46.97 & 5.03 & 2.06 & 1.78 & 49.25 & 38.81 & 5.03 & 1.23 & 0.89 & 40.30 & 50.75 \\
LLMLingua & 6.03 & 1.71 & \textbf{3.33} & 48.48 & 54.55 & 6.03 & 0.81 & 1.58 & 41.79 & 41.79 & 6.03 & 1.50 & \textbf{2.30} & 32.84 & 40.30 \\
LongLLMLingua & 6.03 & 1.71 & 2.90 & 45.45 & 59.09 & 6.03 & 1.48 & 2.00 & 41.79 & 41.79 & 6.03 & 1.50 & 2.11 & \textbf{44.78} & 32.84 \\
LLMLingua-2 & 7.15 & \textbf{5.37} & 2.63 & 57.58 & 48.48 & 7.15 & 1.95 & \textbf{2.30} & 38.81 & \textbf{46.27} & 7.15 & 0.00 & 0.69 & 26.87 & 50.75 \\
FrugalPrompt & 6.00 & 4.24 & 3.06 & 54.55 & 50.00 & 6.00 & 1.46 & 1.81 & 43.28 & 40.30 & 6.00 & 2.88 & 1.07 & 43.28 & 50.75 \\
R2C & 5.71 & 4.60 & 2.60 & 41.30 & 29.50 & 5.71 & 1.10 & \textbf{2.30} & 44.40 & 38.30 & 5.71 & 2.20 & 0.70 & 32.10 & 46.90 \\
PartPrompt & 6.39 & 4.61 & 2.84 & 46.97 & 57.58 & 6.39 & \textbf{2.15} & 1.12 & 41.79 & 40.30 & 6.39 & \textbf{4.27} & 1.07 & 38.81 & \textbf{53.73} \\
Selection-p & 5.96 & 3.08 & 3.19 & \textbf{59.09} & \textbf{68.18} & 5.96 & 1.30 & 2.08 & 37.31 & 38.81 & 5.96 & 1.79 & 0.89 & 35.82 & 52.24 \\
\textbf{Ours} & 6.00 & 0.70 & 1.20 & 39.39 & 46.97 & 6.00 & 1.60 & 1.70 & \textbf{53.73} & 40.30 & 6.00 & 0.90 & 1.50 & 40.30 & \textbf{53.73} \\
\textit{Rank} &  & 9 & 9 & 9 & 8 &  & 4 & 7 & 1 & 6 &  & 8 & 3 & 4 & 2 \\
\bottomrule
\end{tabular}%
}
\caption{Detailed results on LongMemEval across compression ratios. $r$ = actual compression ratio; DA = reference direct token-level F1 (\%); RA = reference reconstruction token-level F1 (\%); DJ = direct LLM-as-judge accuracy (\%); RJ = reconstruction LLM-as-judge accuracy (\%). \textbf{Bold} indicates the best result in each column. The Rank row shows the ranking of our method among all methods for each metric.}
\label{tab:longmemeval-detail}
\end{table*}

\section{Additional Generalization Experiments}
\label{sec:appendix-generalization}

We evaluate the already-evolved compressors on three settings that were not used as evolutionary fitness tasks, without re-evolution or task-specific retuning. The OOD QA evaluation combines 200 MuSiQue and 200 NarrativeQA examples. Because the $1.5\times$ outputs exceed the receiver context budget for these unusually long inputs, we report $2.5\times$, $4\times$, and $6\times$. Long-document summarization, following \citet{cohan-etal-2018-discourse}, and 20 Newsgroups classification~\cite{lang1995newsweeder} each use 1,000 randomly sampled examples and all four target ratios. R2C is not included in the OOD QA or non-QA transfer experiments and is therefore absent from all tables in this section. PartPrompt is included for summarization and classification but omitted from OOD QA because its pipeline is prohibitively slow on ultra-long inputs and frequently fails to reach the requested compression ratio; the OOD comparison therefore uses the remaining established baselines. Tables~\ref{tab:ood-generalization}--\ref{tab:classification-generalization} report the complete Direct and Reconstruction results for these included methods on the 0--1 scale.

\begin{table*}[t]
\centering
\scriptsize
\setlength{\tabcolsep}{2.6pt}
\resizebox{\textwidth}{!}{%
\begin{tabular}{l ccc ccc ccc ccc}
\toprule
& \multicolumn{3}{c}{Direct judge} & \multicolumn{3}{c}{Reconstruction judge}
& \multicolumn{3}{c}{Direct F1} & \multicolumn{3}{c}{Reconstruction F1} \\
\cmidrule(lr){2-4}\cmidrule(lr){5-7}\cmidrule(lr){8-10}\cmidrule(lr){11-13}
Method & 2.5$\times$ & 4$\times$ & 6$\times$
& 2.5$\times$ & 4$\times$ & 6$\times$
& 2.5$\times$ & 4$\times$ & 6$\times$
& 2.5$\times$ & 4$\times$ & 6$\times$ \\
\midrule
FrugalPrompt & .3475 & .2825 & .2475 & .2675 & .2500 & .2000 & .2529 & .2170 & .2043 & .2043 & .1354 & .1252 \\
LLMLingua & .1500 & .1500 & .1125 & .0250 & .0500 & .0375 & .2210 & .1853 & .1974 & .0929 & .0898 & .0926 \\
LLMLingua-2 & .2375 & .1250 & .1625 & .0375 & .0500 & .0750 & .2506 & .2088 & .1752 & .1400 & .1082 & .1187 \\
LongLLMLingua & .1375 & .1500 & .1125 & .0250 & .0375 & .0875 & .2316 & .1866 & .1864 & .0897 & .0948 & .0853 \\
Selection-p & .2000 & .1500 & .1250 & .0750 & .1125 & .0625 & .2332 & .2172 & .2147 & .1394 & .1341 & .1300 \\
Selective-Ctx & .1500 & .2000 & .1125 & .0875 & .0500 & .0500 & .2203 & .2210 & .1551 & .1142 & .1057 & .1054 \\
\textbf{Ours} & \textbf{.3500} & \textbf{.2875} & .2325 & .2500 & .2250 & .1950 & \textbf{.2671} & \textbf{.2395} & \textbf{.2245} & \textbf{.2335} & \textbf{.1509} & \textbf{.1407} \\
\bottomrule
\end{tabular}%
}
\caption{OOD QA transfer on MuSiQue and NarrativeQA without re-evolution or retuning. R2C was not included; PartPrompt is omitted because its ultra-long-input pipeline is prohibitively slow and often fails to reach the requested ratio. The $1.5\times$ setting is omitted because its compressed prompts exceed the receiver context budget.}
\label{tab:ood-generalization}
\end{table*}

\begin{table*}[t]
\centering
\scriptsize
\setlength{\tabcolsep}{2.2pt}
\resizebox{\textwidth}{!}{%
\begin{tabular}{l cccc cccc cccc}
\toprule
& \multicolumn{4}{c}{ROUGE-1} & \multicolumn{4}{c}{ROUGE-2} & \multicolumn{4}{c}{ROUGE-L} \\
\cmidrule(lr){2-5}\cmidrule(lr){6-9}\cmidrule(lr){10-13}
Method & 1.5$\times$ & 2.5$\times$ & 4$\times$ & 6$\times$
& 1.5$\times$ & 2.5$\times$ & 4$\times$ & 6$\times$
& 1.5$\times$ & 2.5$\times$ & 4$\times$ & 6$\times$ \\
\midrule
FrugalPrompt & .4254 & .4168 & .4037 & .3842 & .1419 & .1379 & .1296 & .1141 & .2331 & .2282 & .2238 & .2115 \\
LLMLingua & .3974 & .3603 & .3487 & .3407 & .1096 & .0770 & .0717 & .0671 & .2140 & .1931 & .1896 & .1857 \\
LLMLingua-2 & .4236 & .4082 & .3938 & .3786 & .1337 & .1162 & .1038 & .0931 & .2266 & .2187 & .2096 & .2014 \\
LongLLMLingua & .3941 & .3617 & .3479 & .3391 & .1104 & .0783 & .0730 & .0668 & .2123 & .1944 & .1886 & .1850 \\
PartPrompt & .4134 & .3941 & .3720 & .3594 & .1257 & .1118 & .0926 & .0808 & .2203 & .2109 & .2013 & .1918 \\
Selection-p & .4114 & .3874 & .3696 & .3570 & .1286 & .1087 & .0951 & .0849 & .2247 & .2099 & .2002 & .1924 \\
Selective-Ctx & \textbf{.4260} & .4049 & .3824 & .3590 & .1386 & .1181 & .0993 & .0846 & .2296 & .2175 & .2070 & .1929 \\
\textbf{Ours} & .4216 & .4042 & .3859 & .3712 & .1389 & .1235 & .1027 & .0934 & .2292 & .2215 & .2071 & .1979 \\
\bottomrule
\end{tabular}%
}
\caption{Direct-path summarization results without re-evolution or retuning. R2C was not included in this transfer experiment.}
\label{tab:summarization-direct}
\end{table*}

\begin{table*}[t]
\centering
\scriptsize
\setlength{\tabcolsep}{2.2pt}
\resizebox{\textwidth}{!}{%
\begin{tabular}{l cccc cccc cccc}
\toprule
& \multicolumn{4}{c}{ROUGE-1} & \multicolumn{4}{c}{ROUGE-2} & \multicolumn{4}{c}{ROUGE-L} \\
\cmidrule(lr){2-5}\cmidrule(lr){6-9}\cmidrule(lr){10-13}
Method & 1.5$\times$ & 2.5$\times$ & 4$\times$ & 6$\times$
& 1.5$\times$ & 2.5$\times$ & 4$\times$ & 6$\times$
& 1.5$\times$ & 2.5$\times$ & 4$\times$ & 6$\times$ \\
\midrule
FrugalPrompt & .4185 & .4085 & .3956 & .3797 & .1364 & .1307 & .1214 & .1107 & .2280 & .2196 & .2197 & .2073 \\
LLMLingua & .3909 & .3550 & .3298 & .3250 & .1076 & .0737 & .0574 & .0573 & .2118 & .1921 & .1785 & .1766 \\
LLMLingua-2 & .4090 & .3954 & .3737 & .3649 & .1264 & .1115 & .0916 & .0829 & .2191 & .2092 & .1991 & .1933 \\
LongLLMLingua & .3993 & .3497 & .3268 & .3213 & .1109 & .0696 & .0562 & .0538 & .2133 & .1884 & .1769 & .1735 \\
PartPrompt & .4065 & .3888 & .3721 & .3552 & .1212 & .1053 & .0924 & .0773 & .2183 & .2088 & .1978 & .1912 \\
Selection-p & .3984 & .3839 & .3657 & .3494 & .1200 & .1058 & .0911 & .0789 & .2157 & .2086 & .1958 & .1871 \\
Selective-Ctx & .4111 & .3918 & .3669 & .3487 & .1299 & .1093 & .0899 & .0732 & .2232 & .2091 & .1952 & .1833 \\
\textbf{Ours} & .4091 & .3936 & .3784 & .3493 & .1274 & .1124 & .0878 & .0685 & .2216 & .2123 & .2035 & .1940 \\
\bottomrule
\end{tabular}%
}
\caption{Reconstruction-path summarization results without re-evolution or retuning. R2C was not included in this transfer experiment.}
\label{tab:summarization-reconstruction}
\end{table*}

\begin{table*}[t]
\centering
\scriptsize
\setlength{\tabcolsep}{1.8pt}
\resizebox{\textwidth}{!}{%
\begin{tabular}{l cccc cccc cccc cccc}
\toprule
& \multicolumn{4}{c}{Direct accuracy} & \multicolumn{4}{c}{Reconstruction accuracy}
& \multicolumn{4}{c}{Direct macro-F1} & \multicolumn{4}{c}{Reconstruction macro-F1} \\
\cmidrule(lr){2-5}\cmidrule(lr){6-9}\cmidrule(lr){10-13}\cmidrule(lr){14-17}
Method & 1.5$\times$ & 2.5$\times$ & 4$\times$ & 6$\times$
& 1.5$\times$ & 2.5$\times$ & 4$\times$ & 6$\times$
& 1.5$\times$ & 2.5$\times$ & 4$\times$ & 6$\times$
& 1.5$\times$ & 2.5$\times$ & 4$\times$ & 6$\times$ \\
\midrule
FrugalPrompt & .450 & .430 & .445 & .430 & .475 & .415 & .405 & .410 & .4479 & .4517 & .4601 & .4577 & .4859 & .4164 & .4166 & .4114 \\
LLMLingua & .430 & .470 & .365 & .380 & .470 & .410 & .375 & .335 & .4172 & .4860 & .3832 & .4245 & .4809 & .4157 & .3959 & .3538 \\
LLMLingua-2 & .455 & .420 & .390 & .415 & .480 & .445 & .465 & .440 & .4684 & .4354 & .4024 & .4445 & .4694 & .4568 & .4693 & .4375 \\
LongLLMLingua & .440 & .450 & .380 & .375 & .500 & .450 & .375 & .375 & .4510 & .4598 & .4008 & .4039 & .5163 & .4580 & .4019 & .3614 \\
PartPrompt & .490 & .460 & .460 & .450 & .480 & .470 & .470 & .405 & .4931 & .4754 & .4755 & .4636 & .4908 & .4814 & .4782 & .4186 \\
Selection-p & .425 & .445 & .390 & .370 & .455 & .440 & .390 & .370 & .4408 & .4788 & .4031 & .4223 & .4583 & .4458 & .3845 & .3874 \\
Selective-Ctx & .430 & .420 & .400 & .330 & .470 & .450 & .400 & .390 & .4297 & .4440 & .4456 & .3561 & .4840 & .4516 & .3954 & .3819 \\
\textbf{Ours} & .450 & \textbf{.475} & .435 & .405 & .495 & \textbf{.485} & .455 & .410 & .4843 & \textbf{.4894} & .4530 & .4412 & .5162 & \textbf{.4995} & .4523 & .4166 \\
\bottomrule
\end{tabular}%
}
\caption{20 Newsgroups classification results without re-evolution or retuning. R2C was not included in this transfer experiment.}
\label{tab:classification-generalization}
\end{table*}

\section{Statistical Significance Analysis}
\label{sec:appendix-significance}

The paired-bootstrap analysis retains the original seven-baseline comparison. R2C is reported in the aggregate and per-receiver result tables, but is excluded here because paired per-example outputs aligned with this bootstrap protocol were unavailable; consequently, the ``Best baseline'' and summary counts below do not include R2C.

\begin{table*}[t]
\centering
\small

\begin{tabular}{@{}lllrrrllc@{}}
\toprule
Dataset & Ratio & Best baseline & Best & Ours & $\Delta$ & 95\% CI & $p_{\text{corr}}$ & Outcome \\
\midrule
\multicolumn{9}{l}{\textit{DA: Direct reference accuracy (\%; RACE = exact match, others = token-level F1)}} \\
RACE & 1.5$\times$ & Selective-Ctx & 90.51 & 86.01 & $-$4.50 & [$-$10.3, $+$1.4] & $.285$ & \textsc{Inc.} \\
RACE & 2.5$\times$ & LLMLingua-2 & 87.00 & 69.52 & $-$17.48 & [$-$25.2, $-$10.0] & $<\!.001$ & \textsc{Loss} \\
RACE & 4.0$\times$ & LLMLingua & 78.99 & 66.00 & $-$12.99 & [$-$21.5, $-$4.7] & $.027$ & \textsc{Loss} \\
RACE & 6.0$\times$ & LongLLMLingua & 77.99 & 59.48 & $-$18.51 & [$-$26.9, $-$10.1] & $<\!.001$ & \textsc{Loss} \\
Qasper & 1.5$\times$ & FrugalPrompt & 37.81 & 36.37 & $-$1.44 & [$-$7.4, $+$4.7] & $.722$ & \textsc{Inc.} \\
Qasper & 2.5$\times$ & LLMLingua-2 & 35.35 & 34.90 & $-$0.45 & [$-$6.2, $+$5.3] & $.893$ & \textsc{Inc.} \\
Qasper & 4.0$\times$ & LLMLingua-2 & 30.47 & 29.53 & $-$0.94 & [$-$6.6, $+$4.5] & $.817$ & \textsc{Inc.} \\
Qasper & 6.0$\times$ & LLMLingua-2 & 32.41 & 26.97 & $-$5.44 & [$-$11.0, $+$0.1] & $.131$ & \textsc{Inc.} \\
Multi-doc QA & 1.5$\times$ & LLMLingua-2 & 51.03 & 47.60 & $-$3.43 & [$-$11.5, $+$4.7] & $.711$ & \textsc{Inc.} \\
Multi-doc QA & 2.5$\times$ & FrugalPrompt & 51.39 & 45.67 & $-$5.73 & [$-$14.0, $+$2.5] & $.578$ & \textsc{Inc.} \\
Multi-doc QA & 4.0$\times$ & FrugalPrompt & 44.65 & 39.03 & $-$5.62 & [$-$13.6, $+$2.4] & $.578$ & \textsc{Inc.} \\
Multi-doc QA & 6.0$\times$ & FrugalPrompt & 45.89 & 36.70 & $-$9.19 & [$-$17.3, $-$1.2] & $.276$ & \textsc{Inc.} \\
LongMemEval & 1.5$\times$ & PartPrompt & 4.65 & 2.73 & $-$1.91 & [$-$3.6, $-$0.3] & $.187$ & \textsc{Inc.} \\
LongMemEval & 2.5$\times$ & PartPrompt & 3.88 & 1.37 & $-$2.51 & [$-$4.1, $-$0.9] & $.052$ & \textsc{Inc.} \\
LongMemEval & 4.0$\times$ & PartPrompt & 4.54 & 1.63 & $-$2.91 & [$-$4.7, $-$1.1] & $.052$ & \textsc{Inc.} \\
LongMemEval & 6.0$\times$ & PartPrompt & 3.68 & 1.07 & $-$2.61 & [$-$4.2, $-$1.0] & $.045$ & \textsc{Loss} \\
\midrule
\multicolumn{9}{l}{\textit{RA: Reconstruction reference accuracy (\%; RACE = exact match, others = token-level F1)}} \\
RACE & 1.5$\times$ & LLMLingua & 87.03 & 82.51 & $-$4.52 & [$-$11.1, $+$1.8] & $.285$ & \textsc{Inc.} \\
RACE & 2.5$\times$ & LLMLingua-2 & 75.53 & 72.00 & $-$3.53 & [$-$11.8, $+$4.7] & $.552$ & \textsc{Inc.} \\
RACE & 4.0$\times$ & Selective-Ctx & 70.03 & 56.56 & $-$13.47 & [$-$22.1, $-$4.5] & $.027$ & \textsc{Loss} \\
RACE & 6.0$\times$ & LLMLingua & 64.04 & 56.49 & $-$7.55 & [$-$16.4, $+$1.7] & $.196$ & \textsc{Inc.} \\
Qasper & 1.5$\times$ & FrugalPrompt & 23.92 & 19.10 & $-$4.82 & [$-$9.8, $-$0.0] & $.116$ & \textsc{Inc.} \\
Qasper & 2.5$\times$ & FrugalPrompt & 26.65 & 19.20 & $-$7.45 & [$-$13.0, $-$2.3] & $.024$ & \textsc{Loss} \\
Qasper & 4.0$\times$ & FrugalPrompt & 27.31 & 22.00 & $-$5.31 & [$-$10.6, $+$0.2] & $.139$ & \textsc{Inc.} \\
Qasper & 6.0$\times$ & FrugalPrompt & 27.58 & 19.97 & $-$7.62 & [$-$13.1, $-$2.3] & $.019$ & \textsc{Loss} \\
Multi-doc QA & 1.5$\times$ & FrugalPrompt & 36.04 & 31.03 & $-$5.01 & [$-$12.5, $+$2.6] & $.600$ & \textsc{Inc.} \\
Multi-doc QA & 2.5$\times$ & FrugalPrompt & 41.99 & 33.53 & $-$8.45 & [$-$16.4, $-$0.4] & $.314$ & \textsc{Inc.} \\
Multi-doc QA & 4.0$\times$ & FrugalPrompt & 34.99 & 28.67 & $-$6.33 & [$-$14.0, $+$1.4] & $.490$ & \textsc{Inc.} \\
Multi-doc QA & 6.0$\times$ & PartPrompt & 35.00 & 30.30 & $-$4.70 & [$-$12.5, $+$3.0] & $.600$ & \textsc{Inc.} \\
LongMemEval & 1.5$\times$ & LLMLingua & 2.16 & 1.53 & $-$0.63 & [$-$1.9, $+$0.7] & $.746$ & \textsc{Inc.} \\
LongMemEval & 2.5$\times$ & LLMLingua-2 & 2.61 & 2.20 & $-$0.41 & [$-$1.8, $+$1.1] & $.851$ & \textsc{Inc.} \\
LongMemEval & 4.0$\times$ & LLMLingua-2 & 2.71 & 0.97 & $-$1.75 & [$-$3.1, $-$0.4] & $.122$ & \textsc{Inc.} \\
LongMemEval & 6.0$\times$ & LLMLingua & 2.40 & 1.47 & $-$0.94 & [$-$2.2, $+$0.3] & $.475$ & \textsc{Inc.} \\
\bottomrule
\end{tabular}
\caption{Statistical significance of \textbf{Ours} versus the strongest baseline at each (dataset, ratio) cell on \textbf{reference non-judge accuracy}. $\Delta$ = Ours $-$ Best. 95\% CI from a two-sided bootstrap with 10{,}000 resamples (seed 42). $p_{\text{corr}}$ is Benjamini--Hochberg FDR-corrected within each dataset across all tests. \textsc{Win} = Ours significantly better; \textsc{Loss} = Ours significantly worse; \textsc{Inc.} = statistically inconclusive ($\alpha = 0.05$). An inconclusive result is not evidence of equivalence.}
\label{tab:sig-tokenf1}
\end{table*}

\begin{table*}[t]
\centering
\small

\begin{tabular}{@{}lllrrrllc@{}}
\toprule
Dataset & Ratio & Best baseline & Best & Ours & $\Delta$ & 95\% CI & $p_{\text{corr}}$ & Outcome \\
\midrule
\multicolumn{9}{l}{\textit{DJ: Direct judge accuracy (\%)}} \\
Qasper & 1.5$\times$ & PartPrompt & 61.99 & 62.51 & $+$0.53 & [$-$8.2, $+$9.3] & $.953$ & \textsc{Inc.} \\
Qasper & 2.5$\times$ & LLMLingua-2 & 57.48 & 54.00 & $-$3.48 & [$-$12.8, $+$5.7] & $.650$ & \textsc{Inc.} \\
Qasper & 4.0$\times$ & LLMLingua-2 & 50.51 & 46.48 & $-$4.03 & [$-$13.2, $+$5.2] & $.548$ & \textsc{Inc.} \\
Qasper & 6.0$\times$ & LLMLingua-2 & 47.05 & 39.49 & $-$7.55 & [$-$16.7, $+$1.6] & $.210$ & \textsc{Inc.} \\
Multi-doc QA & 1.5$\times$ & LLMLingua-2 & 60.03 & 62.51 & $+$2.49 & [$-$6.2, $+$11.2] & $.786$ & \textsc{Inc.} \\
Multi-doc QA & 2.5$\times$ & FrugalPrompt & 60.05 & 57.00 & $-$3.05 & [$-$12.4, $+$6.2] & $.750$ & \textsc{Inc.} \\
Multi-doc QA & 4.0$\times$ & LLMLingua-2 & 54.53 & 51.02 & $-$3.51 & [$-$13.0, $+$5.9] & $.731$ & \textsc{Inc.} \\
Multi-doc QA & 6.0$\times$ & FrugalPrompt & 50.09 & 43.02 & $-$7.07 & [$-$16.5, $+$2.2] & $.578$ & \textsc{Inc.} \\
LongMemEval & 1.5$\times$ & Selection-p & 46.01 & 42.94 & $-$3.08 & [$-$12.4, $+$6.2] & $.750$ & \textsc{Inc.} \\
LongMemEval & 2.5$\times$ & Selection-p & 49.03 & 49.48 & $+$0.45 & [$-$8.9, $+$9.7] & $.977$ & \textsc{Inc.} \\
LongMemEval & 4.0$\times$ & PartPrompt & 51.03 & 48.48 & $-$2.56 & [$-$12.0, $+$6.5] & $.835$ & \textsc{Inc.} \\
LongMemEval & 6.0$\times$ & FrugalPrompt & 47.04 & 44.47 & $-$2.56 & [$-$11.8, $+$6.7] & $.816$ & \textsc{Inc.} \\
\midrule
\multicolumn{9}{l}{\textit{RJ: Reconstruction judge accuracy (\%)}} \\
Qasper & 1.5$\times$ & FrugalPrompt & 40.02 & 33.04 & $-$6.98 & [$-$15.7, $+$1.8] & $.223$ & \textsc{Inc.} \\
Qasper & 2.5$\times$ & Selection-p & 42.49 & 28.03 & $-$14.47 & [$-$23.5, $-$5.6] & $.005$ & \textsc{Loss} \\
Qasper & 4.0$\times$ & Selection-p & 40.98 & 32.50 & $-$8.49 & [$-$17.6, $+$0.6] & $.139$ & \textsc{Inc.} \\
Qasper & 6.0$\times$ & FrugalPrompt & 39.47 & 26.51 & $-$12.96 & [$-$21.7, $-$4.2] & $.010$ & \textsc{Loss} \\
Multi-doc QA & 1.5$\times$ & PartPrompt & 40.56 & 33.94 & $-$6.62 & [$-$15.4, $+$2.1] & $.578$ & \textsc{Inc.} \\
Multi-doc QA & 2.5$\times$ & FrugalPrompt & 47.53 & 38.49 & $-$9.04 & [$-$18.1, $+$0.4] & $.424$ & \textsc{Inc.} \\
Multi-doc QA & 4.0$\times$ & LLMLingua-2 & 39.03 & 34.53 & $-$4.50 & [$-$13.6, $+$4.4] & $.708$ & \textsc{Inc.} \\
Multi-doc QA & 6.0$\times$ & FrugalPrompt & 40.06 & 36.48 & $-$3.58 & [$-$12.7, $+$5.6] & $.711$ & \textsc{Inc.} \\
LongMemEval & 1.5$\times$ & PartPrompt & 50.01 & 54.02 & $+$4.01 & [$-$5.4, $+$13.1] & $.750$ & \textsc{Inc.} \\
LongMemEval & 2.5$\times$ & LLMLingua-2 & 54.00 & 52.00 & $-$2.00 & [$-$11.5, $+$7.5] & $.906$ & \textsc{Inc.} \\
LongMemEval & 4.0$\times$ & PartPrompt & 50.52 & 49.54 & $-$0.98 & [$-$9.4, $+$9.7] & $.978$ & \textsc{Inc.} \\
LongMemEval & 6.0$\times$ & Selection-p & 53.08 & 47.00 & $-$6.08 & [$-$15.4, $+$3.4] & $.579$ & \textsc{Inc.} \\
\bottomrule
\end{tabular}
\caption{Statistical significance of \textbf{Ours} versus the strongest baseline at each (dataset, ratio) cell on \textbf{LLM-as-judge accuracy}. $\Delta$ = Ours $-$ Best. 95\% CI from a two-sided bootstrap with 10{,}000 resamples (seed 42). $p_{\text{corr}}$ is Benjamini--Hochberg FDR-corrected within each dataset across all tests. \textsc{Win} = Ours significantly better; \textsc{Loss} = Ours significantly worse; \textsc{Inc.} = statistically inconclusive ($\alpha = 0.05$). An inconclusive result is not evidence of equivalence.}
\label{tab:sig-llmjudge}
\end{table*}

We assess whether the gap between \textbf{Ours} and the strongest of the original seven baselines at each (dataset, ratio, metric) cell is statistically significant. We run a two-sided bootstrap with 10{,}000 resamples (fixed seed 42); the two-sided $p$-value is $2 \min(\Pr[\Delta \leq 0], \Pr[\Delta \geq 0])$ over the bootstrap distribution. $p$-values are corrected with the Benjamini--Hochberg procedure within each dataset across all tests. Tables~\ref{tab:sig-tokenf1} and~\ref{tab:sig-llmjudge} classify each comparison as a statistically significant win, a statistically significant loss, or a statistically inconclusive result. Across the 56 comparisons, 47 are inconclusive, nine are significant losses, and none are significant wins. Broken down by metric, 21 of 24 open-ended token-level F1 comparisons, 22 of 24 judge comparisons, and four of eight RACE exact-match comparisons are inconclusive. These counts describe uncertainty under the present sample size and do not establish performance equivalence.

\section{Compressors' Cost Analysis}
\label{sec:appendix-cost}

\begin{figure}[t]
  \centering
  \includegraphics[width=0.48\textwidth]{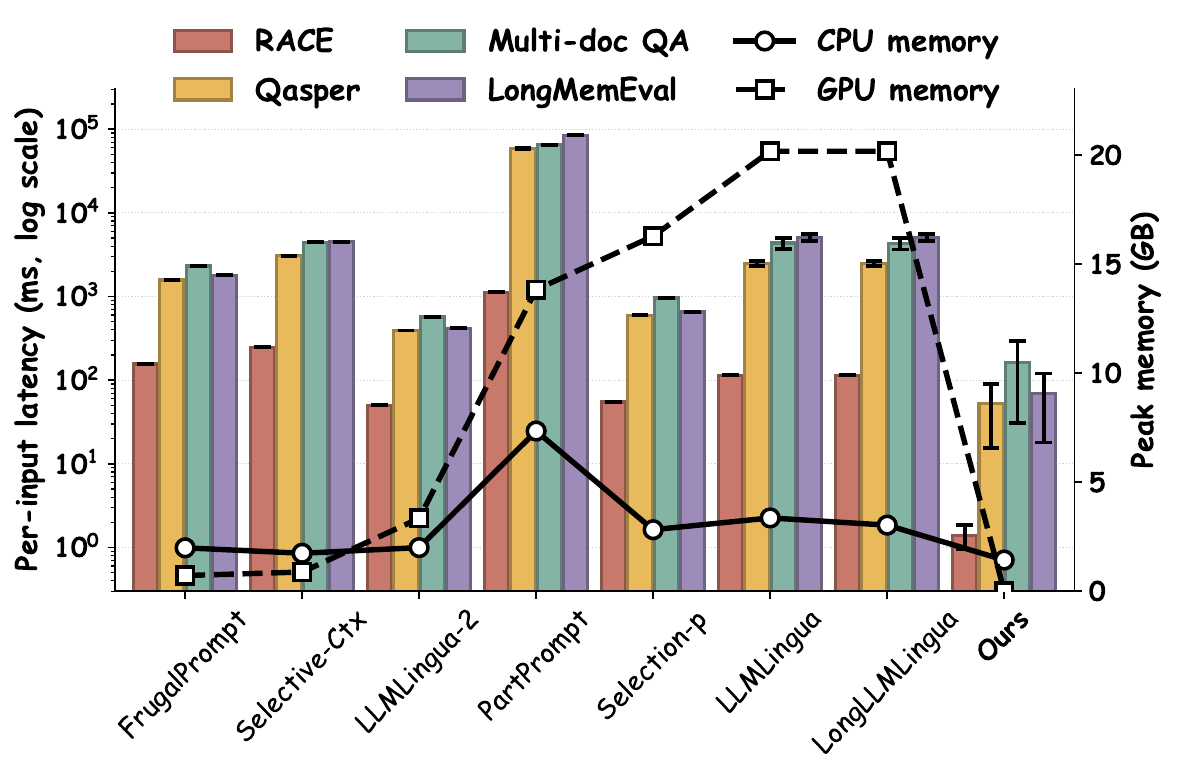}
  \caption{Memory usage and latency across compression methods.}
  \label{fig:cost_figure}
\end{figure}

This appendix reports the latency and memory footprint of our evolved compressor at each of the four target ratios ($1.5\times$, $2.5\times$, $4\times$, $6\times$). Our measurements use the same 40-input subset of the test set (10 inputs per dataset $\times$ 4 datasets). Per-input latency is the median over three repeated calls, and peak CPU memory is the maximum process resident set size observed during the run. Our compressor runs on CPU with only spaCy. Tables~\ref{tab:baseline-cost} and~\ref{tab:baseline-cost-per-dataset-1p5x}--\ref{tab:baseline-cost-per-dataset-6x} compare it with the seven original neural baselines. The baseline measurements use a single NVIDIA A100. Figure~\ref{fig:cost_figure} summarizes the benchmark. 

For the evoluntionary search, it used 120 self-evolution iterations and 126 mutator calls (including failures) for each ratio, totaling about 26 hours and US\$50 around. The resulting four programs use only CPU-side spaCy and deterministic rules at deployment, with no GPU neural scoring or LM forward pass.

\begin{table*}[h]
\centering
\small

\setlength{\tabcolsep}{4pt}
\begin{tabular}{lcccc}
\toprule
\textbf{Dataset} & \textbf{1.5$\times$} & \textbf{2.5$\times$} & \textbf{4$\times$} & \textbf{6$\times$} \\
\midrule
RACE                        & 2.5 \scriptsize{$\pm$1.9}      & 0.7 \scriptsize{$\pm$0.5}     & 1.8 \scriptsize{$\pm$1.2}      & 0.6 \scriptsize{$\pm$0.4} \\
Qasper                      & 164.5 \scriptsize{$\pm$259.1}  & 9.2 \scriptsize{$\pm$7.8}     & 28.8 \scriptsize{$\pm$31.9}    & 9.4 \scriptsize{$\pm$7.7} \\
Multi-doc QA               & 549.5 \scriptsize{$\pm$498.2}  & 13.5 \scriptsize{$\pm$8.5}    & 67.6 \scriptsize{$\pm$53.7}    & 13.5 \scriptsize{$\pm$8.6} \\
LongMemEval                 & 221.6 \scriptsize{$\pm$322.5}  & 10.5 \scriptsize{$\pm$8.9}    & 34.2 \scriptsize{$\pm$36.4}    & 10.2 \scriptsize{$\pm$8.3} \\
\midrule
\textbf{Overall mean}       & \textbf{234.5} \scriptsize{$\pm$370.5} & \textbf{8.4} \scriptsize{$\pm$8.5} & \textbf{33.1} \scriptsize{$\pm$42.0} & \textbf{8.4} \scriptsize{$\pm$8.4} \\
\bottomrule

\end{tabular}
\caption{Per-input compression latency (ms, median per input) of our evolved compressor across the four target ratios. Each cell reports the mean $\pm$ standard deviation over 10 inputs per dataset, with the median of three repetitions taken for each input. Peak process memory is 1.4~GB at every ratio (spaCy \texttt{en\_core\_web\_sm} \textasciitilde12M parameters).}
\label{tab:ours-cost}
\end{table*}

\paragraph{Baseline measurements.}
We measured latency, GPU memory, and CPU memory for the seven original neural baselines using the same harness on a single NVIDIA A100 (40~GB), with one Colab notebook per method and the same input benchmark set described above. Table~\ref{tab:baseline-cost} reports all four target ratios.

\begin{table*}[h]
\centering
\small

\setlength{\tabcolsep}{3pt}
\begin{tabular}{l l c c c c c c c}
\toprule
\textbf{Method} & \textbf{Backbone (params)} & \textbf{Load} & \textbf{GPU peak} & \textbf{CPU peak} & \multicolumn{4}{c}{\textbf{Latency (ms)}} \\
\cmidrule(lr){6-9}
                &                            & \textbf{(s)} & \textbf{(GB)} & \textbf{(GB)} & $1.5\times$ & $2.5\times$ & $4\times$ & $6\times$ \\
\midrule
Selective-Ctx     & GPT-2 (124M)              &  20  &  0.88 & 1.74 & 3087.6 & 3066.6 & 3071.3 & 3049.1 \\
LLMLingua         & Llama-2-7B                &  67  & 20.16 & 3.36 & 3946.1 & 3073.9 & 2630.8 & 2405.2 \\
LongLLMLingua     & Llama-2-7B                &  59  & 20.16 & 3.03 & 3923.6 & 3067.3 & 2628.7 & 2406.1 \\
LLMLingua-2       & XLM-R-large (560M)        &  32  &  3.34 & 2.00 &  359.1 &  357.6 &  357.8 &  356.7 \\
FrugalPrompt      & Electra-large (335M)      &  15  &  0.72 & 1.99 & 1465.0 & 1460.8 & 1459.6 & 1459.8 \\
R2C               & FiD T5-base (220M)        & 9  & 16.89 & 2.39 & 345.4 & 342.9 & 335.4 & 331.7 \\
Selection-p       & Llama-2-7B + adapter      & 111  & 16.27 & 2.82 &  566.7 &  567.6 &  567.0 &  566.8 \\
PartPrompt        & Llama-2-7B + CoreNLP      &  65  & 13.82 & 7.34 & 51784.2 & 53538.8 & $^{*}$52323.0 & 51646.5 \\
\midrule
\textbf{Ours}     & \textbf{spaCy (12M, CPU)} & \textbf{0.6} & \textbf{0.0} & \textbf{1.43} & \textbf{234.5} & \textbf{8.4} & \textbf{33.1} & \textbf{8.4} \\
\bottomrule
\end{tabular}
\caption{Compression cost across methods at all four target ratios. GPU and CPU columns report peak memory. The seven original neural baselines use a single NVIDIA A100.}
\label{tab:baseline-cost}
\end{table*}

\paragraph{Per-dataset breakdown.}
For the original baselines and Ours, the overall numbers in Table~\ref{tab:baseline-cost} aggregate across the 40-input benchmark set; the R2C row instead aggregates the dataset-level values from its audit. Because input length differs by an order of magnitude across the four datasets (RACE \textasciitilde350 words; Qasper \textasciitilde3K; Multi-doc QA \textasciitilde5--6K; LongMemEval exceeding \textasciitilde10K), Tables~\ref{tab:baseline-cost-per-dataset-1p5x}--\ref{tab:baseline-cost-per-dataset-6x} report the per-input latency broken down by dataset, one table per target ratio. Input length substantially affects many neural methods, whereas our latency also varies with the evolved program used at each ratio. $^{*}$ marks the four PartPrompt entries at $4\times$, which are imputed as the mean of the three completed ratios ($1.5$/$2.5$/$6\times$) because the $4\times$ run was interrupted after only five LongMemEval inputs; the imputation is consistent with the overall PartPrompt $4\times$ value in Table~\ref{tab:baseline-cost} and justified because PartPrompt's per-input cost is dominated by Stanford CoreNLP parsing, which is largely independent of target ratio.

\begin{table*}[h]
\centering
\small
\setlength{\tabcolsep}{4pt}
\begin{tabular}{l rrrr}
\toprule
\textbf{Method} & \textbf{RACE} & \textbf{Qasper} & \textbf{Multi-doc QA} & \textbf{LongMemEval} \\
\midrule
Selective-Ctx  & 250.0 & 3088.8 & 4475.3 & 4536.4 \\
LLMLingua      & 114.1 & 3016.4 & 6229.3 & 6424.6 \\
LongLLMLingua  & 114.3 & 2998.1 & 6189.4 & 6392.5 \\
LLMLingua-2    &  51.2 &  391.6 &  573.6 &  420.1 \\
FrugalPrompt   & 156.1 & 1582.7 & 2317.8 & 1803.5 \\
R2C            & 193.3 &  367.8 &  495.8 &  324.8 \\
Selection-p    &  54.5 &  595.7 &  962.1 &  654.3 \\
PartPrompt     & 1125.5 & 57246.2 & 64097.0 & 84667.9 \\
\midrule
\textbf{Ours}  & \textbf{2.5} & \textbf{164.5} & \textbf{549.5} & \textbf{221.6} \\
\bottomrule
\end{tabular}
\caption{Per-input compression latency (ms) at the $1.5\times$ target ratio, broken down by dataset. Values are means over 10 inputs per dataset, with each input represented by the median of three repetitions.}
\label{tab:baseline-cost-per-dataset-1p5x}
\end{table*}

\begin{table*}[h]
\centering
\small
\setlength{\tabcolsep}{4pt}
\begin{tabular}{l rrrr}
\toprule
\textbf{Method} & \textbf{RACE} & \textbf{Qasper} & \textbf{Multi-doc QA} & \textbf{LongMemEval} \\
\midrule
Selective-Ctx  & 247.7 & 3082.5 & 4439.0 & 4497.1 \\
LLMLingua      & 115.1 & 2416.1 & 4544.8 & 5219.7 \\
LongLLMLingua  & 114.2 & 2410.5 & 4518.3 & 5226.3 \\
LLMLingua-2    &  49.4 &  391.9 &  568.6 &  420.7 \\
FrugalPrompt   & 157.7 & 1574.1 & 2316.9 & 1794.6 \\
R2C            & 194.3 &  372.8 &  488.8 &  315.8 \\
Selection-p    &  55.0 &  599.1 &  962.7 &  653.5 \\
PartPrompt     & 1134.9 & 60262.6 & 66352.5 & 86405.3 \\
\midrule
\textbf{Ours}  & \textbf{0.7} & \textbf{9.2} & \textbf{13.5} & \textbf{10.5} \\
\bottomrule
\end{tabular}
\caption{Per-input compression latency (ms) at the $2.5\times$ target ratio, broken down by dataset.}
\label{tab:baseline-cost-per-dataset-2p5x}
\end{table*}

\begin{table*}[h]
\centering
\small
\setlength{\tabcolsep}{4pt}
\begin{tabular}{l rrrr}
\toprule
\textbf{Method} & \textbf{RACE} & \textbf{Qasper} & \textbf{Multi-doc QA} & \textbf{LongMemEval} \\
\midrule
Selective-Ctx  & 248.0 & 3074.0 & 4466.5 & 4496.7 \\
LLMLingua      & 114.7 & 2303.6 & 3605.1 & 4499.6 \\
LongLLMLingua  & 114.3 & 2301.3 & 3575.7 & 4523.3 \\
LLMLingua-2    &  50.6 &  393.1 &  569.1 &  418.3 \\
FrugalPrompt   & 155.6 & 1571.0 & 2314.7 & 1796.9 \\
R2C            & 193.3 &  361.8 &  477.8 &  308.8 \\
Selection-p    &  54.8 &  598.6 &  961.9 &  652.7 \\
PartPrompt     & 1123.8 & 58359.0 & 64891.7 & 84918.0 \\
\midrule
\textbf{Ours}  & \textbf{1.8} & \textbf{28.8} & \textbf{67.6} & \textbf{34.2} \\
\bottomrule
\end{tabular}
\caption{Per-input compression latency (ms) at the $4\times$ target ratio, broken down by dataset. }
\label{tab:baseline-cost-per-dataset-4x}
\end{table*}

\begin{table*}[h]
\centering
\small
\setlength{\tabcolsep}{4pt}
\begin{tabular}{l rrrr}
\toprule
\textbf{Method} & \textbf{RACE} & \textbf{Qasper} & \textbf{Multi-doc QA} & \textbf{LongMemEval} \\
\midrule
Selective-Ctx  & 245.4 & 3061.8 & 4412.2 & 4477.0 \\
LLMLingua      & 114.5 & 2292.6 & 3075.1 & 4138.8 \\
LongLLMLingua  & 114.0 & 2285.6 & 3064.9 & 4159.9 \\
LLMLingua-2    &  49.3 &  391.0 &  568.9 &  417.6 \\
FrugalPrompt   & 155.3 & 1575.0 & 2316.1 & 1792.8 \\
R2C            & 193.3 &  358.8 &  469.8 &  304.8 \\
Selection-p    &  55.1 &  598.7 &  960.9 &  652.4 \\
PartPrompt     & 1111.1 & 57568.2 & 64225.7 & 83680.8 \\
\midrule
\textbf{Ours}  & \textbf{0.6} & \textbf{9.4} & \textbf{13.5} & \textbf{10.2} \\
\bottomrule
\end{tabular}
\caption{Per-input compression latency (ms) at the $6\times$ target ratio, broken down by dataset.}
\label{tab:baseline-cost-per-dataset-6x}
\end{table*}

\section{Best Evolved Compressors}
\label{sec:best_compressor}

This appendix presents the pseudocode of the best evolved compressor at each target ratio. All compressors share a wrapper that enforces the target word count $n_{\text{target}} = \lfloor n_{\text{src}} / r \rceil$ by truncating or back-filling the output. The inner selection logic differs substantially across ratios, reflecting the architectural shift discussed in \S\ref{evolutionary_Analysis}.

\subsection{Best Compressor at \texorpdfstring{$1.5\times$}{1.5x}}

Architecture: \textbf{Flat token-level scoring.} Each token receives a numeric score combining POS category, dependency role, entity chain strength, negation/causal keyword membership, and sentence-level bonuses. The top-$n_{\text{target}}$ tokens are retained in document order.

\begin{mdframed}[backgroundcolor=promptbg, linecolor=promptbg, skipabove=5pt, skipbelow=5pt, innerleftmargin=8pt, innerrightmargin=8pt, innertopmargin=8pt, innerbottommargin=8pt]
\small\ttfamily\raggedright
\textbf{function} compress(text, ratio=1.5):\\
\quad tokens $\leftarrow$ spaCy(text)\\
\quad target\_n $\leftarrow$ round(len(text.split()) / ratio)\\
\quad must $\leftarrow$ causal\_connectives(text) $\cup$ temporal\_markers(text)\\
\quad \\
\quad \textcolor{gray}{\# Sentence-level bonuses}\\
\quad \textbf{for each} sentence s:\\
\qquad bonus $\leftarrow$ 0\\
\qquad \textbf{if} s is first sentence: bonus += 20\\
\qquad \textbf{if} s contains digits: bonus += 18\\
\qquad \textbf{if} s contains entities: bonus += 22\\
\qquad assign bonus to all tokens in s\\
\quad \\
\quad \textcolor{gray}{\# Entity chain strength (dynamic)}\\
\quad \textbf{for each} entity lemma:\\
\qquad strength $\leftarrow$ 40 + min(45, 10 $\times$ occurrence\_count)\\
\quad \\
\quad \textcolor{gray}{\# Token scoring}\\
\quad \textbf{for each} token t:\\
\qquad score $\leftarrow$ 0\\
\qquad \textbf{if} t $\in$ must or t.dep $\in$ \{ROOT, nsubj, dobj, pobj, attr\}: score += 55\\
\qquad \textbf{if} t.dep = neg or t $\in$ keywords: score += 95\\
\qquad \textbf{if} t.ent\_type: score += chain\_strength[t.lemma]\\
\qquad score += POS\_score(t) \textcolor{gray}{\# NUM:88, PROPN:72, NOUN:60, VERB:48/40, ...}\\
\qquad score += sentence\_bonus[t]\\
\quad \\
\quad keep top target\_n tokens by score\\
\quad \textbf{return} kept tokens in document order
\end{mdframed}

\subsection{Best Compressor at \texorpdfstring{$2.5\times$}{2.5x}}

Architecture: \textbf{Tiered fill with hybrid truncation.} The compressor first extracts ROOT verb subtrees as structural skeletons, then fills remaining budget with answer-bearing cues (entities, numbers, negation, connectives) and noun chunk heads. When the collected set exceeds the budget, a hybrid strategy balances priority-based selection with evenly spaced sampling for document coverage.

\begin{mdframed}[backgroundcolor=promptbg, linecolor=promptbg, skipabove=5pt, skipbelow=5pt, innerleftmargin=8pt, innerrightmargin=8pt, innertopmargin=8pt, innerbottommargin=8pt]
\small\ttfamily
\textbf{function} compress(text, ratio=2.5):\\
\quad tokens $\leftarrow$ spaCy(text)\\
\quad target\_n $\leftarrow$ round(len(text.split()) / ratio)\\
\quad keep $\leftarrow$ \{\}\\
\quad \\
\quad \textcolor{gray}{\# Tier 1: ROOT subtree skeletons}\\
\quad \textbf{for each} ROOT verb r:\\
\qquad \textbf{for each} t in subtree(r):\\
\qquad\quad \textbf{if} t.dep $\in$ \{ROOT, nsubj, dobj, attr, ccomp, xcomp, neg\}\\
\qquad\quad\quad or t.ent\_type: add t to keep\\
\quad \\
\quad \textcolor{gray}{\# Tier 2: answer-bearing cues}\\
\quad \textbf{for each} token t:\\
\qquad \textbf{if} t.ent\_type or t.pos = NUM or t.dep = neg\\
\qquad\quad or t $\in$ cue\_words: add t to keep\\
\quad add causal\_connectives(text)\\
\quad add temporal\_markers(text)\\
\quad \\
\quad \textcolor{gray}{\# Tier 3: noun chunk heads}\\
\quad \textbf{for each} noun chunk: add last NOUN/PROPN to keep\\
\quad \\
\quad \textcolor{gray}{\# Hybrid truncation if over budget}\\
\quad \textbf{if} |keep| > target\_n:\\
\qquad priority $\leftarrow$ tokens with entities, capitals, or digits\\
\qquad chosen $\leftarrow$ top half from priority\\
\qquad fill remaining slots by even spacing across keep\\
\quad \\
\quad \textbf{return} kept tokens in document order
\end{mdframed}

\subsection{Best Compressor at \texorpdfstring{$4\times$}{4x}}
Architecture: \textbf{Sentence-then-token cascade.} A two-stage pipeline first scores and selects informative sentences, then applies token-level pruning within the selected sentences if the output still exceeds the budget.

\begin{mdframed}[backgroundcolor=promptbg, linecolor=promptbg, skipabove=5pt, skipbelow=5pt, innerleftmargin=8pt, innerrightmargin=8pt, innertopmargin=8pt, innerbottommargin=8pt]
\small\ttfamily
\textbf{function} compress(text, ratio=4.0):\\
\quad tokens $\leftarrow$ spaCy(text)\\
\quad sents $\leftarrow$ segment(text)\\
\quad target\_n $\leftarrow$ round(len(text.split()) / ratio)\\
\quad \\
\quad \textcolor{gray}{\# Stage 1: Sentence scoring and selection}\\
\quad \textbf{for each} sentence s:\\
\qquad score $\leftarrow$ 0\\
\qquad \textbf{if} s contains entities: score += 2.5\\
\qquad \textbf{if} s contains numbers: score += 2.0\\
\qquad \textbf{if} s contains negation: score += 1.5\\
\qquad \textbf{if} s contains causal connectives: score += 1.0\\
\qquad \textbf{if} s contains temporal markers: score += 1.0\\
\qquad \textbf{if} s contains ROOT verb: score += 1.0\\
\quad select sentences by score until budget filled\\
\quad \\
\quad \textcolor{gray}{\# Stage 2: Token pruning (if still over budget)}\\
\quad \textbf{if} output > target\_n:\\
\qquad keep $\leftarrow$ \{entities, PROPN, NUM\}\\
\qquad add causal \& temporal connectives\\
\qquad \textbf{for each} ROOT verb: add subtree core roles\\
\qquad \textbf{for each} noun chunk: add head NOUN/PROPN\\
\quad \\
\quad \textbf{return} kept tokens in document order
\end{mdframed}

\subsection{Best Compressor at \texorpdfstring{$6\times$}{6x}}
Architecture: \textbf{Sentence selection + subtree deletion + post-processing.} The compressor first selects sentences by surface cues (first sentence, WH-question words, question marks, colons), then removes entire dispensable subtrees and function words within each selected sentence, and finally applies a post-processing chain.

\begin{mdframed}[backgroundcolor=promptbg, linecolor=promptbg, skipabove=5pt, skipbelow=5pt, innerleftmargin=8pt, innerrightmargin=8pt, innertopmargin=8pt, innerbottommargin=8pt]
\small\ttfamily
\textbf{function} compress(text, ratio=6.0):\\
\quad sents $\leftarrow$ segment(text)\\
\quad target\_n $\leftarrow$ round(len(text.split()) / ratio)\\
\quad \\
\quad \textcolor{gray}{\# Stage 1: Sentence selection}\\
\quad kept\_sents $\leftarrow$ \{\}\\
\quad \textbf{for each} sentence s at index i:\\
\qquad \textbf{if} i = 0: add s \textcolor{gray}{\# first sentence always kept}\\
\qquad \textbf{if} s contains \{who,what,when,where,why,how,which\}: add s\\
\qquad \textbf{if} s contains ``?'' or ``:'': add s\\
\quad \\
\quad \textcolor{gray}{\# Stage 2: Subtree deletion + function word removal}\\
\quad \textbf{for each} kept sentence s:\\
\qquad tokens $\leftarrow$ spaCy(s)\\
\qquad \textbf{for each} token t:\\
\qquad\quad \textbf{if} t.dep $\in$ \{advcl, ccomp, xcomp, relcl, acl, appos\}:\\
\qquad\quad\quad drop entire subtree(t)\\
\qquad\quad \textbf{if} t.dep $\in$ \{det, aux, cop, mark, case, punct\}: drop t\\
\qquad\quad \textbf{if} t.pos = ADP and t.lemma $\notin$ common\_preps: drop t\\
\quad \\
\quad \textcolor{gray}{\# Stage 3: Post-processing}\\
\quad output $\leftarrow$ collapse\_repeated\_entities(output)\\
\quad output $\leftarrow$ compress\_enumeration\_blocks(output)\\
\quad output $\leftarrow$\\
\qquad merge\_short\_adjacent\_sentences(output)\\
\quad \\
\quad \textbf{return} output
\end{mdframed}

\section{Resemblance to Existing Compression Methods}
\label{sec:appendix-qualitative}

The evolved programs reuse familiar operations, but their closest analogue changes with the target ratio. At $1.5\times$, token-level pruning resembles Selective Context, LLMLingua, LLMLingua-2, Selection-p, and FrugalPrompt; the distinction is that self-information, perplexity, learned preservation scores, and attention are replaced by explicit linguistic features. At $2.5\times$, the program is closest to PartPrompt because both preserve dependency-based predicate--argument structure. Ours additionally preserves entities, numbers, negation, causal and temporal cues, noun chunks, and document coverage without LM scoring.

At $4\times$, the evolved program is a hybrid that first selects sentences and then prunes within them. At $6\times$, it first keeps the initial sentence and sentences containing WH words, question marks, or colons, resembling rule-based sentence extraction. It then removes subordinate, complement, relative, and appositive subtrees, drops low-content function words, and collapses repeated entities and enumerations. Its outputs therefore consist of compressed propositional skeletons rather than intact extracted sentences; this pruning stage is closest to PartPrompt. We do not claim that each operation is new. Rather, evolution rediscovers useful operations and combines them across linguistic levels and ratios without an LM forward pass at deployment.

\section{Observable Retention and Cases}
\label{sec:appendix-retention-cases}

We audit output behavior at $4\times$ through query-bearing and answer-bearing term retention. Table~\ref{tab:retention-4x} contains the comparative query-term values reported by this audit. Ours retains 72.0\% of query-bearing terms and 70.2\% of answer terms. Its query retention exceeds the four reported LM-scoring baselines even though no explicit query-matching rule survives. Rare-term retention is only mid-range, suggesting that rare entities and values are retained mainly when they belong to a preserved relational structure rather than solely because they are infrequent.

\begin{table*}[t]
\centering
\small
\begin{tabular}{lr}
\toprule
Method & Query-term retention (\%) \\
\midrule
LLMLingua      & 64.4 \\
LongLLMLingua  & 64.6 \\
FrugalPrompt   & 65.5 \\
Selection-p    & 63.4 \\
\textbf{Ours}  & \textbf{72.0} \\
\bottomrule
\end{tabular}
\caption{Available output-level retention audit at $4\times$. The table reports the subset of methods with numerical query-term values; Ours also retains 70.2\% of answer terms.}
\label{tab:retention-4x}
\end{table*}

\paragraph{Success: relation preservation.}
Answering the question about the birthplace of Frank Wartenberg's wife requires two facts: Christiane Wartenberg was born in Prenzlau, and Frank married Christiane in 1977. Ours retains both facts and answers \emph{Prenzlau}. Selective Context retains more rare terms overall but drops Prenzlau and predicts \emph{Siegen}; PartPrompt retains Prenzlau but drops the marriage relation, leaving the receiver unable to connect the answer to the question.

\paragraph{Success: a three-fact chain.}
For the release date of a film starring the actress who played Mia Jones, the answer requires the chain Nina Dobrev $\rightarrow$ \emph{Crash Pad} $\rightarrow$ September 25, 2017. Ours preserves the facts identifying the film and actress and giving the release date. LLMLingua and LongLLMLingua spend much of their budget on earlier unrelated passages; PartPrompt loses the release fact; and LLMLingua-2 retains the date as an isolated span but loses its link to the correct film among competing dates.

\paragraph{Failure: weak structural cues in late passages.}
For a comparison between \emph{A Cafe in Cairo} and \emph{War Drums}, the answer requires linking each film to its director and then comparing the directors' birth years. Ours retains the biographical facts but drops two short, late passages that supply the film--director links, and the receiver returns a director rather than a film title. FrugalPrompt, Selective Context, and Selection-p retain those passages and answer correctly. Thus, contextual or learned salience can be preferable when decisive content is short, appears late, and has weak structural cues in the absence of the query.

These cases also qualify the eliminated-seed analysis. Context-free frequency bonuses and explicit query matching are redundant in the final evolved combination, not generally ineffective. Existing systems replace raw frequency with contextual self-information, perplexity, or learned salience; our programs often recover query relevance indirectly through entities, predicates, dates, and coverage, but the failure case shows where that indirect route breaks.

\section{Additional Comparator Analyses}
\label{sec:appendix-comparator-analyses}

\subsection{TokenShrink Ratio Audit}

To broaden the comparison with non-LM compressors, we additionally evaluated TokenShrink\footnote{\url{https://github.com/chatde/tokenshrink}}, a heuristic compression toolkit that operates without an LM. We therefore ran its default configuration on all 800 test examples. However, it exposes no adjustable target-ratio parameter and left 95.5\% of outputs uncompressed. Its mean actual word-compression ratio was $1.0030\times$, its maximum observed per-example ratio was $1.1266\times$, and none of the 800 examples reached $1.5\times$ (Table~\ref{tab:tokenshrink-audit}). Because it cannot be brought into the paper's $1.5\times$--$6\times$ matched-ratio regime, we report the audit but do not treat TokenShrink as an accuracy baseline at those targets.

\begin{table*}[t]
\centering
\small
\begin{tabular}{@{}lrrrr@{}}
\toprule
Dataset & Unchanged & Avg. ratio & Max. ratio & $\geq1.5\times$ \\
\midrule
RACE         & 94.5\% & 1.0045 & 1.1181 & 0/200 \\
Qasper       & 99.0\% & 1.0006 & 1.0608 & 0/200 \\
Multi-doc QA & 97.5\% & 1.0013 & 1.0587 & 0/200 \\
LongMemEval  & 91.0\% & 1.0056 & 1.1266 & 0/200 \\
\midrule
Overall      & 95.5\% & 1.0030 & 1.1266 & 0/800 \\
\bottomrule
\end{tabular}
\caption{TokenShrink under its default settings. ``Unchanged'' is the fraction of samples left uncompressed; ratios are measured with the paper's common word-count definition.}
\label{tab:tokenshrink-audit}
\end{table*}

\subsection{Scope Relative to ProCut}

ProCut~\cite{xu-etal-2025-procut} estimates attribution over semantic units and prunes low-utility units, whereas PartPrompt applies syntactic-tree pruning and our setting compresses natural-language context documents. ProCut primarily studies structured prompt templates composed of instructions, demonstrations, and heuristic rules. We did not find a public implementation that permitted a faithful matched-ratio evaluation when this experiment was conducted, so we include the methodological comparison but no empirical ProCut row.

\clearpage

\section{Experimental Prompts}
\label{sec:appendix-prompts}

This appendix collects, verbatim, every LLM prompt used in the pipeline: (i)~the system and user prompts that drive the evolutionary mutator, (ii)~the receiver-LLM QA prompts (multiple-choice and open-ended), (iii)~the reconstruction-LLM prompt that expands a compressed string back into prose, and (iv)~the LLM-as-judge prompt used for binary grading of semantic correctness.

Placeholders in curly braces ({\tt \{text\}}, {\tt \{q\}}, {\tt \{compressed\}}, etc.) are substituted at runtime; everything else is sent to the LLM as written.

\subsection{Mutator Prompts}

The mutators are large LLMs that propose edits to a parent compressor's source code via SEARCH/REPLACE diffs. We also use a challenger-round user template that forces the mutator to pursue a specific structural strategy when the population becomes homogeneous. The detailed system prompts are shown in Tables~\ref{tab:mutator1} and~\ref{tab:mutator2}, and the challenger prompt is shown in Table~\ref{tab:challenge}.

\begin{table*}[!htbp]
\centering
\promptcaptionsetup
\begin{promptpanel}{MUTATOR / SYSTEM}{Part 1 of 2}
\begin{Verbatim}[fontsize=\footnotesize,breaklines=true,breakanywhere=true]
You are an expert evolutionary search agent helping discover small Python
`compress(text: str) -> str` functions that reduce document length while
preserving the information a small receiver LM (7-12B) needs to answer questions.

## Research context

This work studies **linguistic-knowledge-based text compression** -- NOT
general "LLM-as-compressor". The paper contribution is demonstrating that
rule-based + structural-linguistic compressors, optimized via evolution,
can match or beat prior LLM-compression baselines (LLMLingua, partprompt,
selective_context) on held-out QA. Therefore:

### LM usage constraints (HARD RULES)

1. **Allowed**: structural linguistic analyzers that produce deterministic
   outputs:
   - `spacy` -- tokens / POS / dependency tree / NER (core tooling)

2. **Allowed (non-LM)**: frequency-based heuristics
   - `wordfreq.zipf_frequency(token)` -- word-frequency table for UID-inspired
     surprisal proxies. NO neural LM.
   - NLTK WordNet -- lexical hierarchies (hypernyms)

3. **FORBIDDEN**:
   - Do NOT prompt an LLM to "compress this text" or "summarize this" --
     that is LLM-as-compressor, which defeats the paper's
     linguistic-knowledge thesis.
   - Do NOT use neural LM surprisal estimators (minicons, GPT-2 perplexity) --
     these are "LM-based scoring" and blur the contribution. Use wordfreq
     instead.
   - Do NOT call the receiver or reconstruction model inside `compress()`.

\end{Verbatim}
\end{promptpanel}
\caption{System prompt for the LLM mutator during evolution}
\label{tab:mutator1}
\end{table*}

\begin{table*}[!htbp]
\centering
\promptcaptionsetup
\begin{promptpanel}{MUTATOR / SYSTEM}{Part 2 of 2}
\begin{Verbatim}[fontsize=\footnotesize,breaklines=true,breakanywhere=true]
## Contract

- You ONLY rewrite code inside `# EVOLVE-BLOCK-START` / `# EVOLVE-BLOCK-END`
  markers. Leave everything outside them byte-identical.
- `compress` must always return a non-empty string.
- The compression ratio `len(text.split()) / len(compress(text).split())`
  must stay in [1.3, 7.0]. Outside this range -> rejected.
- The output can be prose, JSON, PENMAN graph, triples, bullet list -- the
  reconstruction model will expand any format back into prose for QA.
- Fitness = `min(direct_qa_acc, reconstruct_qa_acc)` (after floor gates).
  Floor = 80% x baseline on both paths. Admissibility additionally requires
  n-gram overlap with source in [0.15, 0.75] (prevents fabrication or
  near-identity outputs).

## What good evolution looks like

- Small, targeted edits usually beat full rewrites -- start with the existing
  structure, change one layer at a time.
- Wrap risky logic in `try/except` with a conservative fallback (e.g.,
  first N/3 tokens). A crash is a hard zero on fitness.
- Diversity matters: the archive uses MAP-Elites + behavior fingerprint
  dedup. If your output on 8 fixed probes exactly matches an existing
  archive member, your candidate is rejected as a clone at stage1 (before
  any LLM calls). Tweaking variable names or whitespace doesn't change
  behavior; make real strategy changes.

## Metrics you see above

Every prompt includes the parent's current `comp_ratio`, `direct_qa_acc`,
`reconstruct_qa_acc`, `combined_score`, and which MAP-Elites bin it
occupies. These describe the parent -- you decide how to modify.

\end{Verbatim}
\end{promptpanel}
\caption{System prompt for the LLM mutator during evolution (continued)}
\label{tab:mutator2}
\end{table*}

\begin{table*}[!htbp]
\centering
\promptcaptionsetup
\begin{promptpanel}{MUTATOR / CHALLENGER}{User prompt}
\begin{Verbatim}[fontsize=\footnotesize,breaklines=true,breakanywhere=true]
## CHALLENGER ROUND -- iteration {evolution_round}

The archive contains many similar compressors and the last few iterations
have produced incremental variations. Your task this round is to produce a
**STRUCTURALLY DIFFERENT** compressor using this specific strategy:

**Required strategy:** {forced_strategy}

## Rules

1. Do NOT tweak the parent. Rewrite the entire compress() body.
2. The output behavior MUST match the strategy. Examples:
   - "structured JSON output" -> compress returns `json.dumps({...})`
   - "OpenIE triples" -> compress returns `"subj1|verb1|obj1 | subj2|verb2|obj2"`
   - "AMR PENMAN notation" -> compress returns a PENMAN-formatted graph string
   - "coref collapse" -> compress uses fastcoref cached clusters to
     substitute pronouns with canonical mentions, then drops stopwords
   - "wordfreq filter" -> compress uses `wordfreq.zipf_frequency(token)` to
     drop the most predictable (high-zipf) tokens, keeping informational ones
3. **LM usage constraints** (hard rule, do not violate):
   - You MAY call structural analyzers (spacy, amrlib, fastcoref,
     frame_semantic_transformer, stanza) -- these output deterministic
     linguistic structures
   - You MAY read precomputed caches (`PRECOMPUTED_CACHE`)
   - You MUST NOT directly prompt an LLM to "compress this text" -- that would
     make your approach identical to LLMLingua / LLM-as-compressor
   - You MUST NOT use neural LM surprisal estimators (minicons, GPT-2) -- use
     wordfreq for frequency-based heuristics instead
4. Target compression ratio hint: {target_comp_ratio:.2f}x (but ratio is
   enforced by the V11 wrapper; focus on the strategy, not the ratio)

## Parent context (what NOT to imitate)

Parent compress() body:
{parent_code}

Parent metrics:
{parent_metrics}

## Your task

Write a valid Python function body for `compress(text: str) -> str` that:
(a) implements the required strategy,
(b) produces observably different compressed text than the parent, and
(c) preserves information that allows a small receiver LM to answer
    questions about the original text.

## Output format -- STRICT SEARCH/REPLACE DIFF

You MUST output exactly the SEARCH/REPLACE block format below. The harness
parses this format with a regex; any deviation = your edit is dropped.

<<<<<<< SEARCH
<paste the exact lines of the parent compress() that you want to replace>
=======
<paste the new lines that replace them>
>>>>>>> REPLACE
\end{Verbatim}
\end{promptpanel}
\caption{Challenger prompt for LLM mutators}
\label{tab:challenge}
\end{table*}

\subsection{Receiver LLM Prompts}

The receiver LLM (\texttt{Qwen2.5-7B}, \texttt{Llama-3.1-8B}, \texttt{Gemma-3-12B}) is queried with one of two prompt templates depending on the dataset's answer format. In the Direct path, the receiver LLM reads the compressed text. In the Reconstruction path, a reconstruction LLM first expands the compressed text into prose, after which the receiver LLM reads the reconstructed prose. In both cases, the same receiver template in Table~\ref{tab:receivers} is applied, with {\tt \{text\}} bound to either the compressed string or the reconstructed string.

\begin{table*}[!htbp]
\centering
\promptcaptionsetup
\begin{promptpanel}{RECEIVER / SHARED INSTRUCTION}{All QA tasks}
\begin{Verbatim}[fontsize=\footnotesize,breaklines=true,breakanywhere=true]
IMPORTANT: Output ONLY the requested answer -- no explanations, no chain of
thought, no commentary.

[receiver template follows below]
\end{Verbatim}
\end{promptpanel}
\caption{Prompt for receiver LLMs}
\label{tab:receivers}
\end{table*}

\subsubsection{Multiple-choice QA Prompts}

{\tt \{o\}} in Table~\ref{tab:raceqa} is the option block (e.g.\ \verb|A. ...|, \verb|B. ...|, $\ldots$), and {\tt \{L\}} is the allowed-letter list (e.g.\ \verb|A/B/C/D|) for the multiple-choice questions in the RACE dataset.

\subsubsection{Open-ended QA Prompts}
The remaining three datasets use open-ended answers, so we use the template in Table~\ref{tab:openendQA} as the prompt for receiver LLMs.

\begin{table*}[!htbp]
\centering
\promptcaptionsetup
\begin{promptpanel}{RECEIVER / MULTIPLE-CHOICE QA}{RACE}
\begin{Verbatim}[fontsize=\footnotesize,breaklines=true,breakanywhere=true]
Read the following text carefully.

TEXT:
{text}

QUESTION: {q}
OPTIONS:
{o}

Answer with only the letter ({L}).
\end{Verbatim}
\end{promptpanel}
\caption{RACE dataset QA template}
\label{tab:raceqa}
\end{table*}

\begin{table*}[!htbp]
\centering
\promptcaptionsetup
\begin{promptpanel}{RECEIVER / OPEN-ENDED QA}{Qasper, Multi-doc QA, LongMemEval}
\begin{Verbatim}[fontsize=\footnotesize,breaklines=true,breakanywhere=true]
Read the following passage and answer the question.

PASSAGE:
{text}

QUESTION: {q}

Answer concisely (one phrase or sentence).
\end{Verbatim}
\end{promptpanel}
\caption{Prompt template for open-ended questions}
\label{tab:openendQA}
\end{table*}

\subsection{Reconstruction LLM Prompts}

The reconstruction LLM is drawn from the same three-model pool as the receiver LLMs, but for each sample it differs from the QA receiver; it takes a compressed string and expands it into coherent prose. The reconstructed prose is then fed to a receiver LLM on the Reconstruction path. The detailed prompt is shown in Table~\ref{tab:reconprompt}.

\begin{table*}[!htbp]
\centering
\promptcaptionsetup
\begin{promptpanel}{RECONSTRUCTION MODEL}{Compressed text $\rightarrow$ prose}
\begin{Verbatim}[fontsize=\footnotesize,breaklines=true,breakanywhere=true]
You are a reconstruction model. The compressed input may be in ANY of these
formats: (a) prose text, (b) AMR in PENMAN notation with parentheses and
roles like :ARG0 :ARG1 :location, (c) OpenIE triples formatted 'subj|verb|obj'
separated by '||' or newlines, (d) a bullet list or '- ' fragments, or
(e) telegraphic tokens with connective words dropped. Identify the format,
parse it, and expand into coherent English prose that preserves ALL entities,
numbers, and relationships. Do not add information not in the input.
Output ONLY the reconstructed text, no preamble.

Compressed: {compressed}

Reconstructed:
\end{Verbatim}
\end{promptpanel}
\caption{Prompt for reconstruction LLMs}
\label{tab:reconprompt}
\end{table*}

\subsection{LLM-as-Judge Prompts}

The judge LLM grades each receiver response against the gold answer on a binary scale. Table~\ref{tab:llm_as_judge_prompt} shows the explicit \texttt{1}/\texttt{0} prompt, in which reliability is prioritized and the wrapper preamble suppresses chain-of-thought leakage.

\begin{table*}[!htbp]
\centering
\promptcaptionsetup
\begin{promptpanel}{LLM-AS-JUDGE}{Binary semantic grading}
\begin{Verbatim}[fontsize=\footnotesize,breaklines=true,breakanywhere=true]
You are an expert evaluator. Given a question, a model response, and gold
answer(s), decide whether the response correctly answers the question.

QUESTION: {question}

MODEL RESPONSE: {response}

GOLD ANSWER(S): {golds}

Output ONLY "1" if the response substantively matches at least one gold
answer (paraphrasing OK), or "0" otherwise. No explanation.
\end{Verbatim}
\end{promptpanel}
\caption{LLM-as-judge prompt for GPT-5.4-mini}
\label{tab:llm_as_judge_prompt}
\end{table*}

\end{document}